\documentclass[10pt,conference]{IEEEtran}

\IEEEoverridecommandlockouts
\usepackage{cite}
\usepackage{amsmath,amssymb,amsfonts}
\usepackage{algorithmic}
\usepackage{graphicx}
\usepackage{textcomp}
\usepackage{xcolor}
\def\BibTeX{{\rm B\kern-.05em{\sc i\kern-.025em b}\kern-.08em
    T\kern-.1667em\lower.7ex\hbox{E}\kern-.125emX}}

\usepackage{booktabs} 

\usepackage{lipsum}  
\usepackage{graphicx}
\usepackage{graphics}
\usepackage{multicol}
\usepackage{xspace}
\usepackage{chngpage}
\usepackage{subfig}
\usepackage{multirow}
\usepackage{rotating}
\usepackage{color, colortbl}
\usepackage{algorithm}
\usepackage[font=footnotesize]{caption}

\usepackage{amsthm}

\usepackage{tikz}
\usepackage{collcell}
\usepackage{arydshln}
\usepackage{array}
\usepackage{hhline}

\usepackage{tabularx} 
\usepackage[export]{adjustbox}

\usepackage{pgfplots,pgfplotstable}
\pgfplotsset{compat=1.8,compat/show suggested version=false}
\usetikzlibrary{calc,trees,arrows,patterns,plotmarks,shapes,snakes,er,3d,automata,backgrounds,topaths,decorations.pathmorphing,decorations.markings}
\pgfplotsset{
   /pgfplots/bar  cycle  list/.style={/pgfplots/cycle  list={%
        {black,fill=black!30!white,mark=none},%
        {black,fill=red!30!white,mark=none},%
        {black,fill=green!30!white,mark=none},%
        {black,fill=yellow!30!white,mark=none},%
        {black,fill=brown!30!white,mark=none},%
     }
   },
}

\usepgfplotslibrary{groupplots}

\newtheorem{definition}{Definition}

\newcolumntype{M}[1]{>{\centering\arraybackslash}m{#1}}

\definecolor{color_blue}{RGB}{37,103,185}
\definecolor{color_red}{RGB}{200,0,122}
\definecolor{color_green}{RGB}{80,101,10}

\definecolor{color_orange}{RGB}{0,155,0}
\definecolor{color_brown}{RGB}{160,83,1}
\definecolor{color_pink}{RGB}{18,46,208}
\definecolor{color_purple}{RGB}{158,38,217}
\definecolor{color_navy}{RGB}{190,10,10}

\definecolor{color_black}{RGB}{0, 0, 0}

\usetikzlibrary{external}
\newcommand{\CMPSYSTEM}{VIPER}
\newcommand{\cmpsystem}{VIPER}

\newcommand{\SYSTEM}{BOA}
\newcommand{\system}{BOA}

\newcommand{\ie}{\textit{i}.\textit{e}.}

\newcommand{\figref}[1]{Figure~\ref{fig:#1}}

\newcommand{\secref}[1]{Section~\ref{sec:#1}}

\newcommand{\tblref}[1]{Table~\ref{tbl:#1}}

\newcommand{\algoref}[1]{Algorithm~\ref{algo:#1}}

\newcommand{\eg}{{\em e.g. }}

\newcommand{\demofig}

\def\tabularxcolumn#1{m{#1}}

\newcolumntype{C}[1]{>{\hsize=#1\hsize\centering\small\arraybackslash}X}%
\newcolumntype{D}[1]{>{\hsize=#1\hsize\centering\bfseries\footnotesize\arraybackslash}X}%
\newcolumntype{E}[1]{>{\hsize=#1\hsize\centering\footnotesize\arraybackslash}X}%
\newcolumntype{F}[1]{>{\hsize=#1\hsize\centering\tiny\arraybackslash}X}%

\begin{document}

\title{Comparing and Combining Approximate Computing Frameworks}
\author{\IEEEauthorblockN{Saeid Barati}
\IEEEauthorblockA{\textit{Computer Science Department} \\
\textit{University of Chicago}\\
Chicago, USA \\
saeid@cs.uchicago.edu}
\and
\IEEEauthorblockN{Gordon Kindlmann}
\IEEEauthorblockA{\textit{Computer Science Department} \\
\textit{University of Chicago}\\
Chicago, USA \\
glk@cs.uchicago.edu}
\and
\IEEEauthorblockN{Hank Hoffmann}
\IEEEauthorblockA{\textit{Computer Science Department} \\
\textit{University of Chicago}\\
Chicago, USA \\
hankhoffmann@cs.uchicago.edu}
}

\maketitle

\begin{abstract}



Approximate computing frameworks configure applications so they can
operate at a range of points in an accuracy-performance trade-off
space. Prior work has introduced many frameworks to create approximate
programs.  As approximation frameworks proliferate, it is natural to
ask how they can be \emph{compared} and \emph{combined} to create even
larger, richer trade-off spaces.  We address these questions by
presenting VIPER and BOA.  VIPER compares trade-off spaces induced by
different approximation frameworks by visualizing performance
improvements across the full range of possible accuracies.  BOA is a family of exploration techniques that
quickly locate Pareto-efficient points in the immense trade-off space
produced by the combination of two or more approximation frameworks.
We use VIPER and BOA to compare and combine three different
approximation frameworks from across the system stack, including: one that changes numerical precision, one that skips loop iterations, and one that manipulates existing application parameters.  Compared to simply looking at
Pareto-optimal curves, we find VIPER's visualizations provide a
quicker and more convenient way to determine the best approximation
technique for any accuracy loss.  Compared to a state-of-the-art
evolutionary algorithm, we find that BOA explores 14$\times$ fewer
configurations yet locates 35\% more Pareto-efficient points. 

\end{abstract}

\maketitle

\thispagestyle{plain}
\pagestyle{plain}

\section{Introduction}

Approximation frameworks \emph{configure} applications to operate
within a \emph{trade-off} space where the result accuracy is exchanged for
other benefits, typically increased performance.  Different
approximation frameworks exist across the layers of the system stack.
Some focus on the circuit level
\cite{Palem2003,lingamneni2013,Ingole2015,Chippa2013,Muralidharan2016,Chakrapani2006}.
Others replace expensive hardware with approximations
\cite{samadi2014,Esmaeilzadeh2012,Temam2012,Esmaeilzadeh2006}.  Still
others exist at the programming language and compiler level
\cite{bornholt2014,Kansal2013,Sampson2011,Oh2013,ansel2011,Baek2010,Fousse2007,Kwon2009}.

As approximation methods proliferate, it is natural to question their
interaction; especially:
\begin{itemize}
\item \emph{How to compare the trade-off spaces induced by different
    techniques?} Comparing individual points in the trade-off
  space is easy: simply compare all framework's performance at that point.  Most
  approximation frameworks, however, produce a trade-off space---with
  a range of operating points---so we need techniques that compare
  frameworks across that entire range.
\item \emph{How to combine different techniques' trade-off spaces and
    locate Pareto-efficient points in the new space?} The challenge is
  quickly locating more efficient configurations in the immense combined trade-off space, which is too big
  to search exhaustively.
\end{itemize}

\noindent \textbf{VIPER and BOA: }
To compare approximation frameworks we propose VIPER: Visualizing
Improved PErformance Ratios. While existing techniques use numerical
comparisons \cite{Zhou2011,Holzer2007,Ascia2004,Marti2007,Marti2009}
or simply display Pareto-optimal curves, VIPER produces a visual
representation of the trade-off space.  VIPER produces charts showing
normalized performance for different frameworks across all
possible accuracy loss ranges.  A chart is divided into different,
mathematically meaningful regions that show how much one framework
out-performs others. 

To combine frameworks, we propose BOA: Blending Optimal
Approximations.  BOA is a family of algorithms that locate
Pareto-efficient points in the huge trade-off space produced by
multiple frameworks. In its simplest version, BOA-simple
searches the cross product of Pareto-optimal points from individual
frameworks.  BOA extensions, evaluate more of the search space---either deterministically or probabilistically---including more
near-optimal points.  BOA then returns the
Pareto-efficient points from this search space. 

\noindent \textbf{Summary of Results:}
We consider two case studies.  Both use prior
approximation frameworks from across the system stack: Loop Perforation---a compiler technique (LP)
\cite{Sidiroglou-Douskos2011}, PowerDial---an application-level technique (PD) \cite{Hoffmann2011}, and
the Approximate Math Library---a library that changes numerical precision (AML) \cite{Kwon2009}.  We use eleven applications
covering domains from machine learning to image processing.  Each
application includes multiple inputs that we divide into training and
test sets to evaluate whether combination methods produce
statistically sound results on unseen inputs.

The first case study uses VIPER to compare Loop Perforation,
PowerDial, and the Approximate Math Library.  Loop Perforation simply discards loop iterations with no regard to original
intent. PowerDial , in
contrast, builds off
approximations that already exist in the application; {\it i.e.,}
those envisioned by the original programmer. The Approximate Math Library approximates math functions
({\it e.g.,} $exp$, $log$ and $sqrt$) using variable Taylor series
expansion. While these approximations work at different levels of the
stack, VIPER allows us to quickly compare them and produces more
intuitive visualizations than simply looking at Pareto-optimal curves.

The second case study combines the three approximation techniques and
locates Pareto-efficient points in this new, significantly larger
trade-off space.  We compare BOA to two state-of-the-art design-space
exploration algorithms: Multiple Choice Knapsack Problem (MCKP)
\cite{Yang2003} and Non-Sorting Genetic Algorithm (NSGA-II)
\cite{Deb2002}.  Compared to these two approaches, BOA achieves:
\begin{itemize}

    
\item \textbf{More efficient configurations}: \system{}-simple produces
  48.2\% and 35.1\% more Pareto-efficient points than MCKP and
  NSGA-II, respectively. 

\item \textbf{More reliable behavior on unseen inputs}: \system{}-simple
  finds statistically meaningful Pareto-efficient configurations that are not sensitive
  to input data and are more likely to be efficient on
  an unseen set of inputs. That is, the correlation between \system{}'s behavior on training (seen) and test (unseen) inputs is much higher than MCKP and NSGA-II. 
\end{itemize}
Somewhat surprisingly, BOA produces much better results while using a
much simpler search technique than the prior works to which it is
compared.  The fact that simpler search methods can produce better
results is a key contribution of this work.  The primary insight is
that, empirically, we find that optimal combinations of approximation
frameworks tend to involve configurations that are near-optimal for
the individual frameworks.  Therefore, BOA explores combinations derived from these points with high probability.  In
contrast, MCKP does not consider enough non-optimal configurations and
gets stuck in local minima, while NSGA-II explores too many
non-optimal configurations---avoiding the worst local minima, but also
stopping short of the true optimal combinations.  Thus, BOA's method
represents a compromise that works well for approximation frameworks,
whose optimal combinations tend to be near the individually optimal
points.

\noindent \textbf{Contributions:}
\label{sec:contributions}
\begin{itemize}
\item Introduction of VIPER to visually compare approximation
  trade-off spaces over their entire range.
\item Comparison analysis---based on VIPER---of the three approximation
  frameworks (Loop Perforation \cite{Sidiroglou-Douskos2011},
  PowerDial\cite{Hoffmann2011} and Approximate Math
  Library\cite{Kwon2009}).
\item Proposal of variations of BOA for fast locating Pareto-efficient configurations in
  combined trade-off space.
\item Open-source release of the VIPER and BOA tools. 
\end{itemize}


\section{Motivation and Background}
\label{sec:background}


\subsection{Approximation Across the System Stack}
Approximation frameworks reduce runtime (or energy) by allowing
output quality degradation. Hardware approximation computes inexactly in return for
reduced energy, area, or time
\cite{lingamneni2013,Chippa2013,Esmaeilzadeh2011}.  Many software
approximation techniques allow specific
software components to be replaced by approximate variants; \eg{} 
skipping loop iterations or replacing
math operations with Taylor-series expansions
\cite{Omer,Patterns2010,ICSE2010,Hoffmann2009,Samadi2013,Sidiroglou-Douskos2011,Hoffmann2011,Park2016,Kwon2009,Mathar2009,Abad2015,JouleGuard,Fousse2007,Canino2018}.
Some approaches use machine learning
to replace exact computation with a faster, less accurate learned variant
\cite{Esmaeilzadeh2012,Temam2012,Muralidharan2016,Sui2016,ALERT1,ALERT2}.
Languages support approximation allowing
specification of variants for key functionality and formal
analysis of their effects \cite{Sorber2007,Baek2010,Sampson2011,ansel2011,Kansal2013,bornholt2014,Proteus,LAB,Ent}.
Other mechanisms guarantee that
approximate programs will maintain some quality or energy guarantees
either through program analysis \cite{Ringenburg2015,Carbin2013,Darulova2013} or runtimes with formally analyzable dynamic adaptation \cite{JouleGuard,CoAdapt,FSE2017,FSE2015,Meantime,TAAS2017,ASPLOS2018,AdaptCap}.

\subsection{Comparing Approximation Frameworks}
Our
intuition is that some approximation frameworks will produce better results (e.g.,
higher performance for the same accuracy) than others in different
situations. Hence, we need a method to compare frameworks across their
full range of accuracy and choose the best one for a specific usage
scenario.
Points in these \emph{trade-off spaces} correspond to
\emph{configurations} of the approximation framework.  Point-by-point
comparison is unfeasible since trade-off spaces include numerous
configurations, many of which are not useful. Typically, only the
Pareto-optimal points are used for comparison.
 
\begin{figure}[t]
\begin{center}
\input{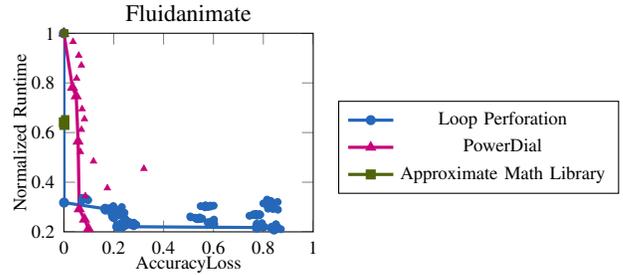}%
\caption{\texttt{fluidanimate}'s accuracy/runtime trade-off space
  with Loop Perforation (LP), PowerDial (PD), and Approximate Math Library (AML).  The closer to origin, the
  more efficient configuration. }
\label{fig:motivationFA}
\end{center}
\end{figure}

As an example, we compare three approximation frameworks: PowerDial
\cite{Hoffmann2011}, Loop Perforation \cite{Hoffmann2009,ICSE2010,Sidiroglou-Douskos2011},
and the Approximate Math Library \cite{Kwon2009}.  We pick these three
frameworks because (1) they are either easily recreated or publicly
available, requiring no specialized language or hardware support, and
yet (2) they are representative of approaches applied at different
levels of the system stack.  PowerDial is an application-level
approach that exploits existing trade-offs envisioned by the
application developers.  Loop Perforation creates approximate
applications by applying a compiler transformation to selectively skip
loop iterations. The Approximate Math Library changes computation, and
while implemented in software, it is a good proxy for approximation
techniques that change hardware arithmetic units.
\figref{motivationFA} illustrates the trade-off spaces induced by
these three approaches for the \texttt{fluidanimate} benchmark.  Each
point represents the normalized runtime and accuracy loss.
 
Comparison of approximation frameworks across their full range of
accuracy is necessary, as not all users have the same accuracy
requirements.  We find, however, that looking at Pareto-optimal curves
like those in \figref{motivationFA}) is unsatisfying and
rarely makes it obvious which approximation method is better across a
range of operating points.  Moreover, while for a certain range of
accuracy one framework might perform better, another framework might
produce higher performance at different accuracy ranges. 
This motivates VIPER, a tool that allows users to tell---at
a glance---which framework has the best performance for any range of
accuracy loss.  

\subsection{Combining Approximation Frameworks}
\figref{motivationFA} suggests that none of the three approximation
frameworks is uniformly best. Furthermore, the fact
that the three are broadly representative of approaches from different
levels of the ssystem stack motivates us to combine them for better 
accuracy/runtime tradeoffs than any individual framework.  
The challenge is that merging multiple
frameworks leads to an enormous trade-off space, which is infeasible
to explore exhaustively.  Table \ref{tbl:exploredConfigs} lists the
number of points in the trade-off spaces of Loop Perforation,
PowerDial, and Approximate Math Library for sample benchmarks.  
For example, the \texttt{x264} benchmark takes up to 4 weeks to test
all combined configurations with a single input.  Considering multiple
inputs should be tested for statistically sound results, the
unfeasibility of exhaustive exploration is obvious.

More formally, combining approximation frameworks requires
quickly locating Pareto-efficient configurations in the new, larger
trade-off space. Exploring large trade-off spaces is well-studied and 
has produced
two broad classes of approach. The first is carefully selecting 
and exhaustively searching a subset of the combined trade-off space 
\cite{Givargis2001,Yang2003}. The second class intelligently
traverses the entire combined trade-off space---not limiting the
initial configuration combinations, but exploreing a small number of the
total \cite{Ascia2004,Palermo2009,Zitzler2012}.  Among
these intelligent search techniques, NSGA-II---a genetic
algorithm-based approach---has repeatedly out-performed other proposals \cite{Deb2002,Zitzler2012}.


While prior work has proven effective for application-specific
processor design, we find that it is not the best match for combining
approximation frameworks.  Specifically, heuristic exploration of genetic algorithms appear to
causes two issues: (1) in an effort to avoid local minima, they produce less
efficient combinations (see Section
\ref{sec:coverageMetricEvaluation}) and (2) they add too much randomization 
that leads to lower correlation between training and
test inputs (see Section \ref{sec:inputSensitivity}).

\section{Comparing Approximation Frameworks}
\label{sec:compare1}


\subsection{Terminology}
To produce performance/accuracy tradeoffs, any
approximation framework must have one or more tunable
\emph{parameters}.  The values assigned to the parameter set represent
a \emph{configuration} and the range of possible parameter settings is
a \emph{configuration space}.  Each configuration represents a
\emph{trade-off} between performance and accuracy.  The
\emph{trade-off space} (or \emph{design space}) is the set of all
possible trade-offs; {\it i.e.,} the range of achievable performance
and accuracy.  

We consider large search spaces and often do not
know the true optimal values for which we are searching.  We therefore
distinguish between \emph{Pareto-optimal}---meaning we know
that a point is on the true Pareto-optimal frontier----and 
\emph{Pareto-efficient}---meaning a point on the  estimated, 
 unknown Pareto-optimal frontier.  Thus, if we say a point is 
Pareto-efficient, it is better than all
other points found so far, but we do not know that it is truly
Pareto-optimal.


\subsection{Numerical Comparisons}
\label{sec:coverageanddominance}
For large trade-off spaces, a point-by-point
comparison is not possible.  Therefore, prior work 
has introduced analytical methods
for comparing trade-off spaces based on the number of
Pareto-optimal---if the trade-off space is known---or
Pareto-efficient---if the trade-off space is estimated---points 
from each framework.

A point in our accuracy-performance trade-off space is a 2D vector with
$runtime$ and $accuracyLoss$. Ideally, we would have zero
run time and zero accuracy loss; {\em i.e.,} instantaneously get a
perfect answer, leading to:
\theoremstyle{definition}
\begin{definition}{Objective Function:}
  Given points $x_{1}$ and $x_{2}$ in the objective
  to be minimized is $f(x)$ where:
\begin{equation}
\begin{gathered}
f(x_1) < f(x_2)  \iff accuracyLoss(x_1) < accuracyLoss(x_2) \\ 
                 \& \enspace runtime(x_1) < runtime(x_2)
\end{gathered}
\end{equation}
\end{definition}

\begin{figure}[tb]
\centering
  \includegraphics[width=0.35\textwidth]{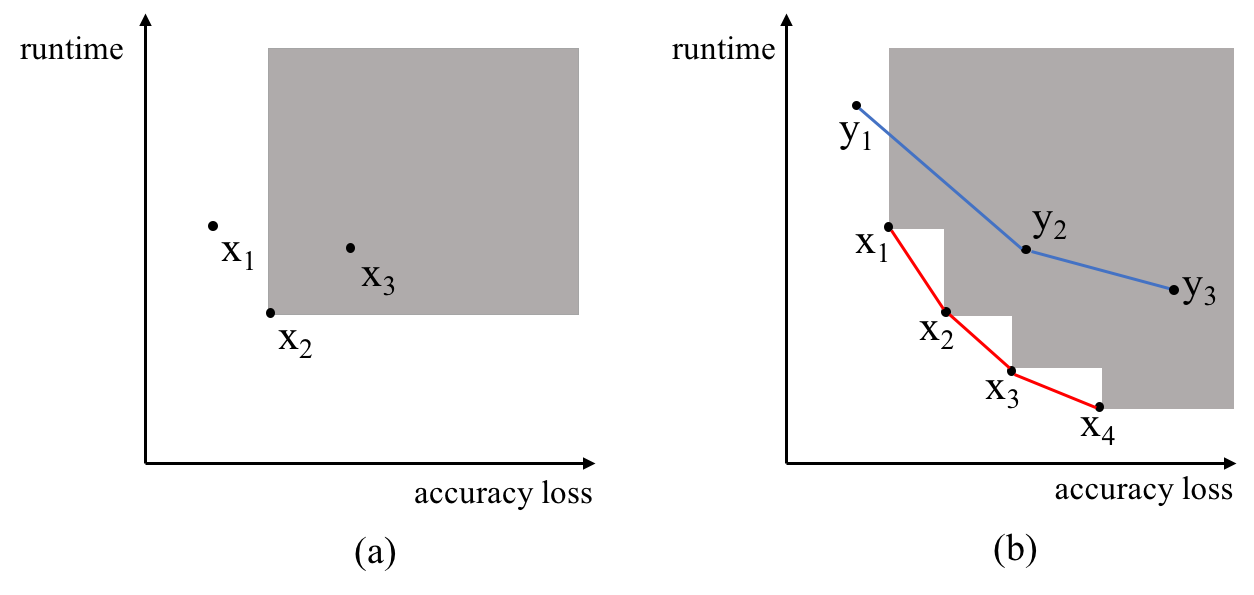}
  \caption{Dominance (a) and Coverage (b) Functions (from
      \cite{Palermo2009}).  In (a) $x_2$ dominates $x_3$ as $x_2$ is
    both faster and more accurate.  In (b), $X$ covers $2/3$ of $Y$
    because $y_2$ and $y_3$ are dominated by at least one point in
    $X$. }
\label{fig:coverageScenarios}
\end{figure}

Points closer to the origin represent more
efficient configurations. Given the objective function $f(x)$, we
determine if a point is more efficient than another by:
\theoremstyle{definition}
\begin{definition}{Dominance:}
  Given points $x_1$ and $x_2$, we say:
\begin{equation}
\begin{gathered}
x_1 \succeq x_2 \textnormal{ (weakly dominates) if } f(x_1) \leqslant f(x_2) \\
x_1 \succ x_2 \textnormal{ (dominates) if } f(x_1) < f(x_2) 
\end{gathered}
\end{equation}
\end{definition}
A point is Pareto-optimal if it is not dominated by any other point. A
point is Pareto-efficient if we do not know of another point that
dominates it. \figref{coverageScenarios}(a) illustrates an example of
dominance where point $x_3$ is dominated by point $x_2$.

\emph{Coverage} quantifies the number of Pareto-efficient points 
 produced by different techniques \cite{Palermo2009}:
\theoremstyle{definition}
\begin{definition}{\emph{Coverage}}
  is the dominance ratio of the Pareto-efficient curves induced by two
  separate frameworks.  If $X$ and $Y$ are two Pareto-efficient
  curves, and $x$ and $y$ represent points on them respectively, then:
\begin{equation}
C(X,Y)=\frac{| \{\: y \in Y \: | \: \exists x \in X : x \succeq y \: \} |}{|Y|}
\end{equation}
\end{definition}
$C(X,Y)=1$ means that all points in Y are weakly dominated by points
in X; {\it i.e.,} all points of $X$ provide lower runtime for the same
accuracy loss than the points of $Y$.

\figref{coverageScenarios}(b) illustrates the coverage of curve $X$
with respect to curve $Y$. The point $y_2$ and $y_3$ on the $Y$ curve
are dominated by at least one point on the curve $X$---$x_2$, for
example---therefore $C(X,Y)=\frac{2}{3}$. In contrast, no point on $X$
is dominated by a point on curve $Y$ which means $C(Y,X)=0$. By this
metric, we consider $X$ more efficient, but note that the curve $Y$
extends through a larger range within the trade-off space; {\em i.e.,}
$y_1$ is a useful point which neither
dominates nor is dominated by any points in $X$.  The coverage
function is non-symmetric ($C(X,Y) \;\neq\; C(Y,X)$) and usually their
sum does not equal 1 \cite{Holzer2007}. Hence, we need a metric that
considers both coverage functions simultaneously:
\theoremstyle{definition}
\begin{definition}{\em Difference Of Coverage}
  compares coverage for two different Pareto-efficient curves.
\begin{equation}
DOC(X,Y)=C(X,Y)-C(Y,X)
\end{equation}
\end{definition}

As a result, when $DOC(X,Y)\geq 0$, that fraction of $Y$ points which
are dominated by $X$ is greater than $X$ points that are dominated by
$Y$. Higher $DOC$ implies one set is more efficient than the other. 
If $DOC(X,Y)$ is close to zero both may provide the same efficiency. 
This metric is widely used by in multi-objective optimization problems 
 \cite{Ascia2004,Marti2007,Marti2009}.

\subsection{VIPER}
\label{sec:compare}
Numerical comparisons suffer from two
major shortcomings. First, they do not show the full range of
accuracy loss induced by each framework. Second, as seen
in \figref{motivationFA}, the best approximation framework varies as
accuracy loss changes. Numerical metrics---like DOC---have
limited expressiveness; \figref{coverageScenarios}(b) shows that $y_1$
is a useful point, but DOC makes $X$ look uniformly better than $Y$.

As an alternative to numerical methods, researchers have used Pareto-optimal
 curves to compare frameworks, but this graphical evaluation has 
proven problematic \cite{Givargis2001,Knowles1999}. 
While curves may look compact, they
can be different by orders of magnitude; \eg{} when there is a
steep slope and a large range covered, a small change in
one dimension (\eg accuracy loss) leads to a significant shift in the
other (\eg runtime). 

\newcommand{\LineComment}[1]{\hfill $\triangleright$ \scriptsize{#1}}

\begin{algorithm}[th]
  \caption{\cmpsystem{}.}
  \begin{algorithmic}[1]
    \footnotesize \REQUIRE $M,B$ \LineComment{Lower convex hulls for framework $M$ and baseline $B$}
    \STATE $MinX = Max(Min(M.x), Min(B.x))$ \LineComment{lower bound}
    \STATE $MaxX = Min(Max(M.x), Max(B.x))$ \LineComment{upper bound}
    \STATE $step = (MaxX-MinX)/1000$ 

  \FOR {$accuLoss =  MinX; accuLoss<MaxX;accuLoss+=step$}
        \STATE $M_i \leftarrow$ find point on $M$ where $M_{i}.x < accuLoss < M_{i+1}.x$
        \STATE $B_j \leftarrow$ find point on $B$ where $B_{j}.x < accuLoss < B_{j+1}.x$ 
    \STATE $\hat{y}_{M} \leftarrow$ interpolate runtime between $M_i$ and $M_{i+1}$ where $x=accuLoss$
    \STATE $\hat{y}_{B} \leftarrow$ interpolate runtime between $B_j$ and $B_{j+1}$ where $x=accuLoss$  
    \STATE $perfImprovRatio[accuLoss]=\hat{y}_{M} / \hat{y}_{B}$
        
  \ENDFOR  
    
    \STATE NORMALIZE($perfImprovRatio$) \LineComment{limit the ratio to [0,1]}  
  \newline
    \RETURN $perfImprovRatio$ \LineComment{array of points} 
  \end{algorithmic}
  \label{algo:perfImprovRatio}
\end{algorithm}

To provide an alternative visualization of approximation frameworks 
we introduce \CMPSYSTEM{}, which illustrates the relative performance 
of frameworks 
 for any range of accuracy loss.
\algoref{perfImprovRatio} explains how \CMPSYSTEM{} calculates the
\emph{performance improvement ratio} (PIR) of one framework $M$ over a
baseline $B$.  The PIR is the ratio of the frameworks' 
performance at a given accuracy loss.  
To make the charts readable, PIR is in 
the range [0,1].  PIR$=1$ means a configuration is the fastest in the 
space, while PIR$=0$ is the slowest.  A configuration with PIR$=0.6$ 
achieves 60\% of the maximum speedup.

First, \CMPSYSTEM{} finds the lower and upper bounds for the accuracy
loss which defines the range of comparison. Then, this range gets
divided by a parameterized granularity. We use a granularity of 1000
in this paper as larger values produced no benefit and
smaller values make the charts less clear. For each $accuLoss$ value,
\CMPSYSTEM{} finds the corresponding runtime in both frameworks.
Afterwards, we search for the nearest points on each lower convex hull
where their accuracy loss is smaller than $accuLoss$ (identified with
$M_i$ and $B_j$ points respectively). Then, we interpolate the runtime
of the specified $accuLoss$ for both frameworks (named as
$\hat{y}_{M}$ and $\hat{y}_{B}$), and divide these interpolated
$runtime$ values to compute the performance improvement ratio.
Finally, we normalize the ratio to [0,1]. When we
compare more than two frameworks, the ratio is normalized to the 
lowest and highest among all.


\cmpsystem{} then charts the PIR across the
range of accuracy loss.  The values for the baseline $B$ will be a
straight horizontal line.  Values of $M$ above that line indicate that
$M$ achieves higher performance for that accuracy loss.  If the line
for $M$ stays above that for $B$ for a greater range of accuracy loss,
it means $M$ has found more efficient configurations, on average.  The
color shading on the plot background indicates the highest performance
method for that accuracy loss from the multiple frameworks. Therefore,
if the plot's background is dominated by a single color, the
corresponding method provides the more efficient configurations. 
Thus, \cmpsystem{} allows
users to see at a (literal) glance, whether one approximation
framework is clearly superior to another.

\section{Case Study 1: Comparing Frameworks}
\label{sec:compareAFEvaluation}

\subsection{Experimental Setup}
\label{sec:experimentalsetup}
We use a dual socket Intel Xeon E5 server system with 20 physical
cores at 2.9 GHz, hyperthreading, and 32 GB memory.  Table
\ref{tbl:benchmarks} lists the used benchmarks, from Parsec 3.0
\cite{bienia11benchmarking2011} and Rodinia 3.1 \cite{Che2009}.  Table
\ref{tbl:benchmarks} also contains the description, type, application
accuracy metric, and default run-time for each benchmark.  Accuracy
loss is the error relative to the most accurate configuration. Shorter 
runtime interprets as higher performance. 
\texttt{blackscholes}'s only tunable parameter is the number of prices
to estimate and modifying it does not affect accuracy.  Thus,
PowerDial has no effect on \texttt{blackscholes}.  Similarly,
\texttt{canneal}, \texttt{heartwall}, \texttt{kmeans}, and
\texttt{x264} use math functions infrequently; the Approximate Math
Library is not applicable to them.

We evaluate each suite across multiple inputs and compare the median
across the inputs for this evaluation. In this section, we are
evaluating known frameworks, thus we use the training inputs from Table
\ref{tbl:benchmarks}. In subsequent sections---where we present new
techniques---we divide inputs into training and test and build
combinations of frameworks using the training data and then use the
test data to ensure our selected combination works well on previously
unseen data.

\def\tabularxcolumn#1{m{#1}}

\newcolumntype{C}[1]{>{\hsize=#1\hsize\centering\small\arraybackslash}X}%
\newcolumntype{D}[1]{>{\hsize=#1\hsize\centering\bfseries\footnotesize\arraybackslash}X}%
\newcolumntype{E}[1]{>{\hsize=#1\hsize\centering\footnotesize\arraybackslash}X}%
\newcolumntype{F}[1]{>{\hsize=#1\hsize\centering\tiny\arraybackslash}X}%
\newcolumntype{G}[1]{>{\hsize=#1\hsize\centering\scriptsize\arraybackslash}X}%
\newcolumntype{H}[1]{>{\hsize=#1\hsize\centering\bfseries\scriptsize\arraybackslash}X}%

\begin{table*}[tb]
  \centering
  \caption{Benchmarks used for evaluation.}
\footnotesize
\begin{tabularx}{\textwidth}{|D{0.6}|E{1.4}|E{1.35}|E{1.35}|E{0.3}|}

\hline 
\cellcolor[gray]{0.8} {Benchmarks} & \cellcolor[gray]{0.8}{\textbf{Accuracy Metric}} & \cellcolor[gray]{0.8} {\textbf{Training inputs}} & \cellcolor[gray]{0.8}{\textbf{Test Inputs}} & \cellcolor[gray]{0.8}{\textbf{Runtime (sec)}} \\ 
\hline 
Blackscholes &  Average Relative Error of Prices  & 30 lists with 1M initial prices & 90 lists with 1M initial prices & 3.2 \\ 
\hline 
Bodytrack & Average Distance of Poses & sequence of 100 frames & sequence of 261 frames & 3.1 \\ 
\hline 
Canneal &  Average Relative Routing Cost & 30 netlists with 400K+ elements & 90 netlists with 400K+ elements & 6.88 \\ 
\hline 
Fluidanimate &  Distance between Particles & 5 fluids with 100K+ particles & 15 fluids with 500K+ particles & 17.2 \\ 
\hline 
Heartwall &  Average Relative Error of heart frames & sequence of 30 ultrasound images & sequence of 100 ultrasound images & 11.6 \\ 
\hline 
Kmeans  &  Distance between Cluster Centers & 30 vectors with 256K data points & 90 vectors with 256K data points & 3.1 \\ 
\hline 
Particlefilter & Distance between Particles  &  sequence of 60 frames & sequence of 240 frames & 12.9 \\ 
\hline 
Srad  & Image Diff (RMSE) & 3 images with 2560*1920 pixels & 9 images with 2560*1920 pixels & 22.6 \\ 
\hline 
Streamcluster &  Distance between Cluster Centers & 3 streams of 19k-100K data points & 9 streams of 100K data points &  30 \\ 
\hline 
Swaptions  & Average Relative Error of Prices  & 40 swaptions & 160 swaptions & 6.2 \\ 
\hline 
x264 & Relative  PSNR+Bitrate & 4 HD videos of 200+ frames  & 12 HD videos of 200+ frames & 7.7 \\ 
\hline 
\end{tabularx} 
\label{tbl:benchmarks}
\end{table*}


We evaluate three approximation frameworks.  PowerDial (PD) transforms
an application's command line parameters into \textit{software knobs}
that are automatically manipulated to trade accuracy for performance
\cite{Hoffmann2011}.  Each application has tunable \emph{knobs}, which
can take different values, and an assignment of values to knobs is a
configuration. Loop Perforation (LP) identifies \textit{perforatable
  loops} whose iterations can be skipped to produce faster, but less
accurate results \cite{Sidiroglou-Douskos2011}.  A set of loops and
perforation rates is a configuration.  The Approximate Math Library
(AML) substitutes math functions with a variable Taylor series
expansion.  A set of functions and their number of terms is the
configuration.


\subsection{Comparison by Difference of Coverage}
\label{sec:compareAFDOC}
To compare Loop Perforation and PowerDial, we calculate the average
coverage function ($C(X,Y)$ from Eq. 3) across all benchmarks
for both.  
On average, Loop Perforation covers
only $0.3664$ Pareto-optimal points of PowerDial, while PowerDial
covers $0.4416$ Pareto-optimal points of Loop Perforation. Thus,
$DOC(LoopPerforation, PowerDial) = -0.075$ which shows the slight
superiority of PowerDial over Loop Perforation, on average. 
On the other hand,
negative values of $DOC(AML, PD)=-0.9135$ and $DOC(AML, LP)=-0.929$ 
shows significant inferiority of Approximate Math
Library against other frameworks, on average.

\def\tabularxcolumn#1{m{#1}}

\newcolumntype{C}[1]{>{\hsize=#1\hsize\centering\small\arraybackslash}X}%
\newcolumntype{D}[1]{>{\hsize=#1\hsize\centering\bfseries\footnotesize\arraybackslash}X}%
\newcolumntype{E}[1]{>{\hsize=#1\hsize\centering\footnotesize\arraybackslash}X}%
\newcolumntype{F}[1]{>{\hsize=#1\hsize\centering\tiny\arraybackslash}X}%

\subsection{Comparison by Pareto-optimal Curves}
\label{sec:compareAFParetoOptimalCurves}

\captionsetup[subfigure]{labelformat=empty}

\begin{figure*}[t]
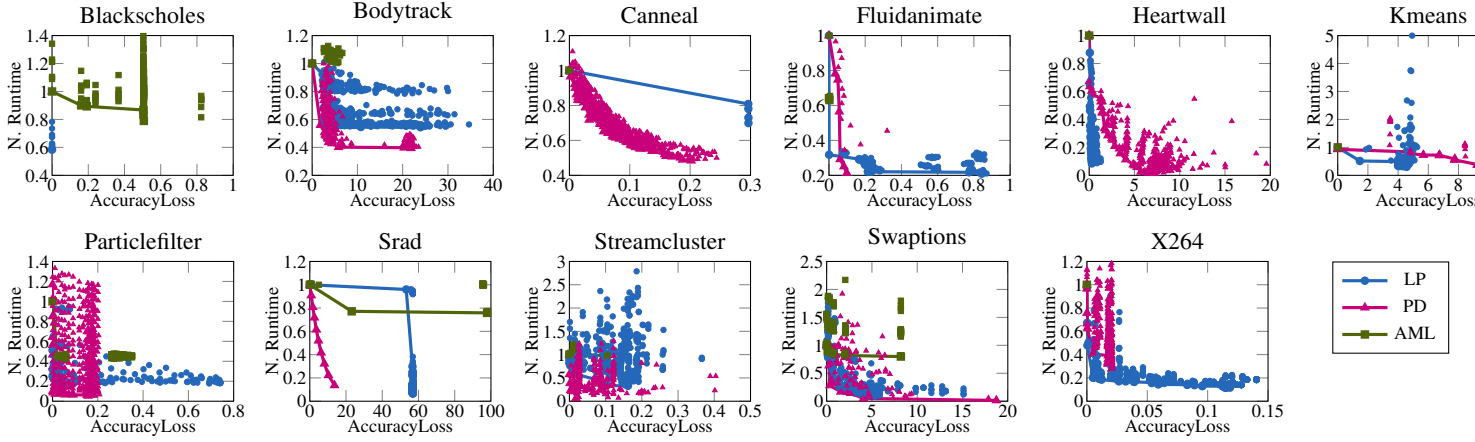
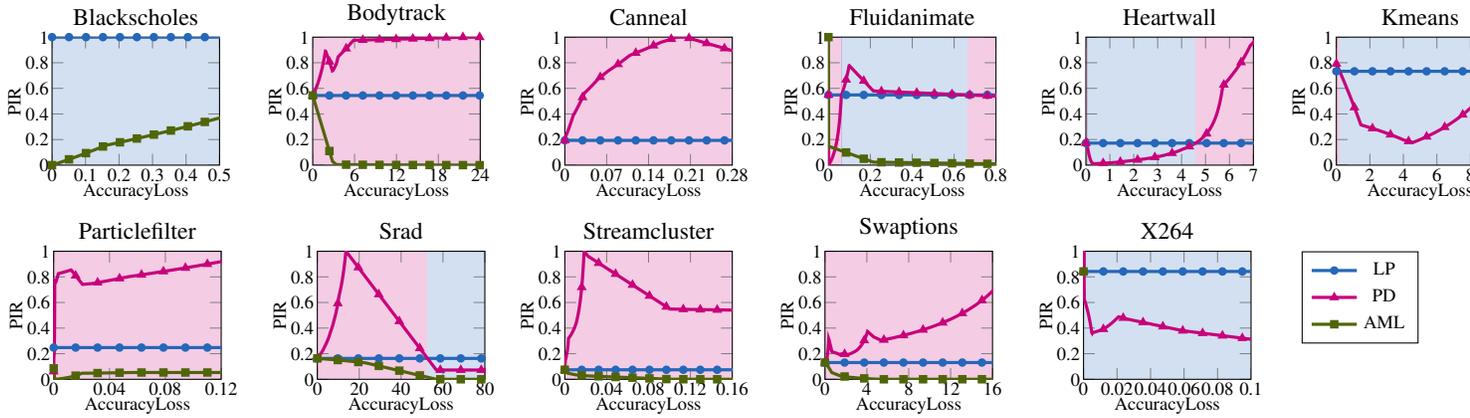

\newcolumntype{V}{>{\centering\arraybackslash} m{0.1\linewidth} }
\newcolumntype{C}[1]{>{\hsize=#1\hsize\centerline\arraybackslash}X}%

\centering

\subfloat[(a) Performance/AccuracyLoss trade-off spaces for PowerDial, Loop Perforation, and Approximate Math Library.]{\label{fig:cs1configSpace}
\hskip0.0cm \begin{tabular*}{\textwidth}{c @{\extracolsep{\fill}} c@{\extracolsep{\fill}} c@{\extracolsep{\fill}} c@{\extracolsep{\fill}} c@{\extracolsep{\fill}} c@{\extracolsep{\fill}}}
\subfloat{\input{img/cs1configSpace/Blackscholes-configSpace.tex}} &
\subfloat{\hspace*{3pt}\input{img/cs1configSpace/Bodytrack-configSpace.tex}} &
\subfloat{\hspace*{3pt}\input{img/cs1configSpace/Canneal-configSpace.tex}} &
\subfloat{\hspace*{3pt}\input{img/cs1configSpace/Fluidanimate-configSpace.tex}} &
\subfloat{\hspace*{-55pt}\input{img/cs1configSpace/Heartwall-configSpace.tex}} &
\subfloat{\hspace*{-55pt}\input{img/cs1configSpace/Kmeans-configSpace.tex}} \\[-3ex] 
\subfloat{\hspace*{3pt}\input{img/cs1configSpace/Particlefilter-configSpace.tex}} &
\subfloat{\hspace*{3pt}\input{img/cs1configSpace/Srad-configSpace.tex}} &
\subfloat{\hspace*{3pt}\input{img/cs1configSpace/Streamcluster-configSpace.tex}} &
\subfloat{\hspace*{3pt}\input{img/cs1configSpace/Swaptions-configSpace.tex}} &
\subfloat{\hspace*{5pt}\input{img/cs1configSpace/X264-configSpace.tex}} \\
\end{tabular*}
}

\vspace{-1.5em}

\subfloat[(b) VIPER comparison of PowerDial, Loop Perforation, and Approximate Math Library.]{\label{fig:cs1viperLines}
\hskip0.0cm \begin{tabular*}{\textwidth}{c @{\extracolsep{\fill}} c@{\extracolsep{\fill}} c@{\extracolsep{\fill}} c@{\extracolsep{\fill}} c@{\extracolsep{\fill}} c@{\extracolsep{\fill}}}
\subfloat{\input{img/cs1viperLines/Blackscholes-cs1viperLine.tex}} &
\subfloat{\hspace*{3pt}\input{img/cs1viperLines/Bodytrack-cs1viperLine.tex}} &
\subfloat{\hspace*{3pt}\input{img/cs1viperLines/Canneal-cs1viperLine.tex}} &
\subfloat{\hspace*{3pt}\input{img/cs1viperLines/Fluidanimate-cs1viperLine.tex}} &
\subfloat{\hspace*{-55pt}\input{img/cs1viperLines/Heartwall-cs1viperLine.tex}} &
\subfloat{\hspace*{-55pt}\input{img/cs1viperLines/Kmeans-cs1viperLine.tex}} \\[-3ex] 
\subfloat{\hspace*{3pt}\input{img/cs1viperLines/Particlefilter-cs1viperLine.tex}} &
\subfloat{\hspace*{3pt}\input{img/cs1viperLines/Srad-cs1viperLine.tex}} &
\subfloat{\hspace*{3pt}\input{img/cs1viperLines/Streamcluster-cs1viperLine.tex}} &
\subfloat{\hspace*{3pt}\input{img/cs1viperLines/Swaptions-cs1viperLine.tex}} &
\subfloat{\hspace*{5pt}\input{img/cs1viperLines/X264-cs1viperLine.tex}} \\
\end{tabular*}
}
\caption{Comparison of PowerDial, Loop Perforation, and Approximate
  Math Library.  (a) shows Pareto-optimal frontiers and (b) shows
  VIPER.}
\label{fig:comparecs1}
\end{figure*}

\figref{cs1configSpace} illustrates the frameworks' trade-off spaces.
Each point represents a configuration.  The y-axis is runtime
normalized to the default configuration and the x-axis is the accuracy
loss. Each framework's Pareto-optimal curve is shown in the same color
as the trade-off space. These plots highlight how configurations cover
wide ranges of runtime and accuracy loss.  While in some cases---\eg{}
\texttt{canneal}, \texttt{kmeans}, and \texttt{srad}---Pareto-optimal
curves are easily to compare, in other benchmarks---\eg{}
\texttt{particlefilter} and \texttt{swaptions}---comparison is
unfeasible.  For \texttt{fluidanimate}, the Pareto-optimal curves
intersect multiple times; the best approximation framework differs
across the range of accuracy loss.

\subsection{Comparison by \cmpsystem{}}
\label{sec:compareAFPIR}

Just viewing the Pareto-optimal curves in \figref{cs1configSpace}
provides limited intuition as
differences are not always visible. We use \cmpsystem{} to
compare these frameworks in Figure \textcolor{red}{3b}.  The y-axis
represents the performance improvement ratio (PIR), while the x-axis
illustrates accuracy loss. The horizontal line represents Loop
Perforation: points above that line mean the corresponding technique
is faster than Loop Perforation. The backdrop color indicates the 
best method for an accuracy loss.

VIPER shows only useful configurations---if one dominates others, 
VIPER shows a small range of accuracyLoss.  
Thus, some ranges are not shown in the Figure 
3b plots comparing to accuracy loss ranges in 
\figref{cs1configSpace} because there is no benefit 
to increasing $accuracyLoss$.

\cmpsystem{} provides the following insights:
\begin{itemize}
\item It illustrates how frameworks perform within a
  specific accuracy loss range. For instance, while PowerDial finds a
  higher performance configuration than Loop Perforation for
  \texttt{bodytrack} and \texttt{canneal}, its performance is
  worse for \texttt{kmeans} and \texttt{x264} for most accuracies.
\item While the distinction between frameworks is not clear in
  \figref{cs1configSpace} for \texttt{streamcluster} and
  \texttt{swaptions}, \cmpsystem{} allows quick, obvious comparison. 
\item \cmpsystem{} clearly illustrates the intersection of Pareto-optimal 
  curves; \eg{}, in  \texttt{fluidanimate} and \texttt{srad}.
\end{itemize}


Since \cmpsystem{} only requires trade-off spaces to compare, it can
be applied to any approximation frameworks regardless of
system level.  We believe \cmpsystem{} provides clear 
insights, which are instantly visually recognizable and 
mathematically meaningful.
VIPER is not a replacement for existing methods, but a 
complement that simplifies comparison. 

\section{Combining Approximation Frameworks}
\label{sec:combine}
The prior section shows that none of Loop Perforation, PowerDial, or
the Approximate Math Library is uniformly best. This observation
motivates us to combine frameworks.  At one level, this process is
quite easy---just create a new trade-off space that is
the cross product of all configurations in the original frameworks.
The challenge, of course, is quickly locating the Pareto-efficient
points in the resulting massive combined trade-off space.

We meet this challenge with the \system{} family of search
algorithms.  All \system{} methods select a subset of the combined
trade-off space and exhaustively search that subspace space.  
The first algorithm,
\system{}-simple, only considers configurations in the cross-product
of individual frameworks' Pareto-optimal configurations.  This
technique produces a relatively small set of points to search, but may
be subject to local minima if the approximation
frameworks are not independent.

Unfortunately, most approximation frameworks are not independent.  
For example, Loop Perforation changes the number of loop
iterations within an application; PowerDial may change convergence
criteria.  When we combine configurations from these frameworks, we
find that some configurations that were Pareto-optimal when
considering only the original frameworks are now far from optimal.
Conversely, we empirically find that some configurations that were not
Pareto-optimal in the original frameworks combine to be Pareto-optimal
when we consider multiple frameworks together. 
These observations motivate us to expand
\system{}-simple to include more non Pareto-optimal 
configurations in combination.

\system{}-flex expands the combined search space to 
consider the configurations that produce a trade-off 
within a user-defined threshold
of Pareto-optimal.  This technique searches more points and tends to
find more efficient combinations, but it is deterministic.  A common
way to avoid local minima in large search spaces is expanding the 
exploration area with some form
of randomization.  We follow this approach with the last algorithm:
\system{}-prob, which probabilistically selects
configurations from each individual framework to combine.
Specifically, it uses a sigmoid probability function, so that the
closer points are to Pareto-optimal, the more likely they are to be
included in the combined trade-off space.  \system{}-prob includes
most of the same points as the other frameworks, but includes some
outliers with small probability, making it more robust in the presence
of local minima.  

\noindent \textbf{BOA-simple:}
The simple version of \system{} forms the cross-product of all
Pareto-optimal configurations from the individual frameworks. 
After executing on evaluation platform, \system{}-simple returns the 
Pareto-efficient configurations found in this
combined trade-off space. 
The worst case complexity of \system{}-simple is bounded by
$O(m+log(m))*2^{\frac{n}{m}}$ where $m$ is the number of frameworks to
be merged and $n$ contains the total number of parameters of all
approximation frameworks\cite{Givargis2001}.\footnote{For the purpose of time
  complexity analysis, we assume each approximation knob can take on
  two values only, however, in reality, parameters may be assigned a
  larger number of values.}  In our experiments, input parameters are
the sum of the number of loop rates, software knobs, and Taylor series
bounds that represent our three frameworks. 
While the algorithm has exponential
complexity, it is practical because so few configurations lie on the
Pareto-optimal frontiers produced by individual frameworks (see
\figref{comparecs1} and Table \ref{tbl:distributedConfigs}).

\noindent \textbf{BOA-flex:}
\system{}-flex augments \system{}-simple with a user-specified
selection threshold, as shown in
\algoref{paretoThreshold}. This threshold also removes some
inconsistency that may arise due to experimental noise; {\it i.e.,} it
is possible that for high variance applications, the true
Pareto-optimal configurations cannot be found with confidence, so
adding the threshold makes the search more robust.   Specifically,
\system{}-flex considers all configurations whose trade-off is within
the user-specified threshold of a Pareto-optimal trade-off. 

This threshold is specified in terms of \emph{normalized Euclidean
  distance}.  All trade-offs are normalized so
that accuracy loss and runtime range from 0 to 1.  
Accuracy loss of 1 means the lowest
quality.  
A runtime of 1 is the slowest execution time. 
A trade-off point is the output of executing a configuration and, 
is thus, a pair of accuracy loss and runtime. Having
normalized all configurations accuracy loss and runtime, we can then
compute the Euclidean distance between the trade-offs of two separate
configurations.  Given this definition, the threshold specifies how
close to Pareto-optimal a trade-off must be for it to be included in
the search. For example, the threshold is $0.05$, and then the
algorithm will include any configuration whose accuracy loss/runtime
trade-off is within 5\% of a Pareto-optimal point. If the threshold is
zero, this algorithm is equivalent to \system{}-simple.

\begin{algorithm}[th]
  \caption{\system{}-flex: expands search space by $threshold$.}
  \begin{algorithmic}[1]
    \footnotesize \REQUIRE $frameworks$  \LineComment{trade-off spaces of frameworks}
    \footnotesize \REQUIRE $threshold$ \LineComment{User-defined $threshold$}
    \STATE $Combination$ = [] \LineComment{configurations to explore}
    \FOR{$f$ in $frameworks$} 
      \STATE ${Pareto-opt}_f \leftarrow$ Get-Pareto-Opt($f$)
      \FOR{Config $C_i$ in $Pareto-opt$}
        \FOR{Config $C_j$ in $f$}
            \IF {$NormalizedEuclideanDistance(C_i,C_j)$ $\leqslant$ $threshold$} 
                \STATE $Combination$.append($C_j$)  
            \ENDIF
        \ENDFOR
      \ENDFOR 
    \ENDFOR 
    \newline 
    \RETURN $Combination$ \LineComment{Set of points to explore} \newline
  \end{algorithmic}
  \label{algo:paretoThreshold}
\end{algorithm}

\noindent \textbf{BOA-prob:}
While \system{}-flex expands the combined search space, it only
considers additional configurations that are close to an individual
framework's Pareto-optimal curve. To make \system{} even more robust
to local minima, \system{}-prob employs a sigmoid probability function
to include a few points that are further from the individual
frameworks' Pareto-optimal curves:
\begin{equation} \label{eq:5}
\begin{gathered}
S(C_j) = \frac {1} {1 +exp(\frac {-\Delta + \beta}{\gamma})}
\end{gathered}
\end{equation}
where $\Delta$ is the normalized Euclidean distance between
configuration $C_j$ and the nearest Pareto-optimal configuration.
$\beta$ is the horizontal shift and $\gamma$ decides the curve's smoothness. 
\algoref{paretoRandom} shows how \system{}-prob uses Equation \ref{eq:5}.

\begin{algorithm}[th]
  \caption{\system{}-prob: probabilistic search space expansion.}
  \begin{algorithmic}[1]
    \footnotesize \REQUIRE $frameworks$  \LineComment{trade-off spaces of frameworks}
    \STATE $Combination$ = [] \LineComment{Pareto-efficient configurations}
    \FOR{$f$ in $frameworks$} 
      \STATE ${Pareto-opt}_f \leftarrow$ Get-Pareto-Opt($f$)
      \FOR{Config $C_i$ in $Pareto-opt$}
        \FOR{Config $C_j$ in $f$} 
          \STATE $\Delta$ = $NormalizedEuclideanDistance$($C_i$,$C_j$) 
          \STATE $S({C_j})$ = $\frac{1}{1 +exp(\frac{-\Delta + \beta}{\gamma})}$   
          \STATE r=rand() \LineComment{Random number between 0 and 1}
          \IF {($r$ < $S({C_j})$) or ($\Delta d=0$)} 
              \STATE $Combination$.append($C_j$)  
          \ENDIF
        \ENDFOR
      \ENDFOR
    \ENDFOR 
    \newline
    \RETURN $Combination$ \LineComment{Set of points to explore} \newline
  \end{algorithmic}
  \label{algo:paretoRandom}
\end{algorithm}

We use constants $\beta=0.2$ and $\gamma=0.01$ so there is a 92\%
chance of including the point that has $\Delta < 0.05$, and 50\%
chance of selecting the point with $\Delta < 0.1$.  At $\Delta=0.2$,
there is less than 1\% chance of including the point in the
combination. If $\Delta=0$, it means that $C_j$ is actually
Pareto-optimal and \system{}-prob includes it. The interdependent
parameters $\beta$ and $\gamma$ control the size of combined
trade-off space and the exploration time.

\section{Case Study 2: BOA Evaluation}
\label{sec:Evaluation}

We compare variations of \system{} to prior exploration techniques 
using the same
experimental setup from 
\secref{experimentalsetup}.  We now split our inputs
into training and test data sets as shown in  Table \ref{tbl:benchmarks}.
For each exploration
technique, we first use the training inputs to find Pareto-efficient
configurations, then we evaluate those points using the separate test data.


\subsection{Points of Comparison}
We compare \system{} to state-of-the-art approaches for
locating Pareto-efficient points in large trade-off spaces:
\begin{itemize}
\item \textbf{MCKP}: The \emph{multiple choice knapsack problem}
  variant of the classic knapsack problem has classes of items and
  must choose one item from each class.  MCKP has been used to find
  Pareto-efficient processor designs in the performance-power space
  for application specific embedded processors \cite{Yang2003}.  We
  declare each framework to be a class.  MCKP then selects the
  Pareto-optimal configurations of each class while keeping the
  default values for other classes.  This creates a new, small
  trade-off space which can be searched with brute force.  
\item \textbf{NSGA-II}: The non-dominated sorting-based
  multi-objective evolutionary algorithm (NSGA-II) explores large
  trade-off spaces to find Pareto-optimal configurations using an
  evolutionary genetic algorithm \cite{Deb2002}. NSGA-II is the
  state-of-the-art for multiobjective optimization of embedded
  processors that navigate performance-power trade-offs, having been
  cited over 20,000 times as  it finds better configurations in
  less time than other evolutionary algorithms 
  \cite{Zitzler2012,Knowles1999}.
\end{itemize}


\subsection{Comparison by Difference of Coverage}
\label{sec:coverageMetricEvaluation}
\begin{figure}[tb]
\begin{center}
	\input{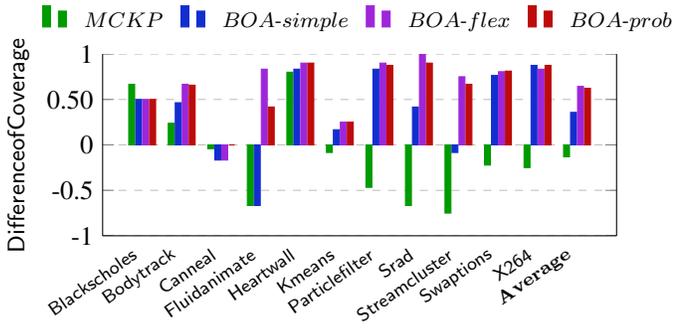}
  \caption{Difference of coverage over NSGA-II. Higher bars are
    better.}
  \label{fig:convergence}
\end{center}
\end{figure}

Recall from Section \ref{sec:compare1} that difference of coverage
(DOC) implies the efficiency of one curve over another. 
\figref{convergence} displays the difference of
coverage for various techniques over NSGA-II per benchmark and on
average. The y-axis shows $DOC(X, NSGA)$ which is DOC of exploration
technique $X$ over NSGA-II (see \secref{coverageanddominance}).
Negative values of $DOC(X, NSGA)$ indicate that $X$ does not find as
many Pareto-efficient points as NSGA-II. Conversely, positive values
imply that technique $X$ provides that many more values that dominate
NSGA-II. \system{}-flex and \system{}-prob on average locate
52.8-65.6\% more Pareto-efficient configurations than NSGA-II. 
\system{}'s superiority is due to its focus on the configurations 
that have been shown to be Pareto-optimal on individual frameworks. 
NSGA-II starts the exploration from a random set of points in the 
combined trade-off space and iteratively looks for more efficient points. 
MCKP uses the individual Pareto-optimal curves but keeps the rest of 
frameworks at default configurations. Since the frameworks are not fully 
independent, we empirically find the some configurations that were not 
Pareto-optimal in the original frameworks become part of a Pareto-efficient 
curve of combined trade-off space when we consider multiple frameworks 
together. By expanding \system{} with threshold and probabilistic 
exploration, we search more points, resulting in more efficient 
configurations. In short, MCKP does not search enough combinations, 
while NSGA-II searches too many.  By restricting the search to points 
likely to be near the Pareto-optimal frontier for individual frameworks, 
BOA achieves the right balance and the best empirical results.
\emph{This data shows that, for approximate computations, \system{}
  produces many more efficient configurations than prior
  state-of-the-art search techniques.}

\subsection{Comparison by VIPER}
\label{sec:perfImprovmentEvaluation}

\captionsetup[subfigure]{labelformat=empty}
\begin{figure*}[pt]
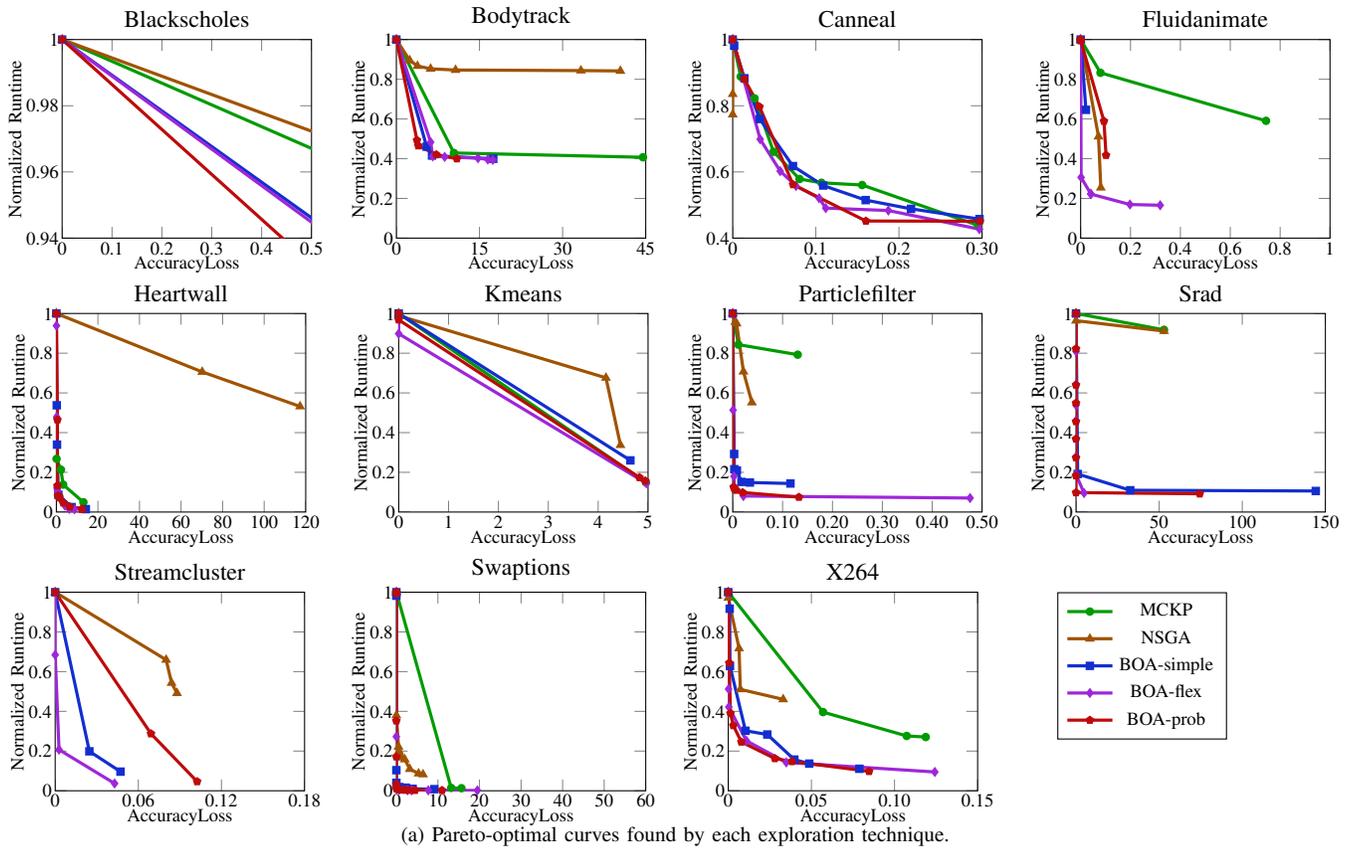
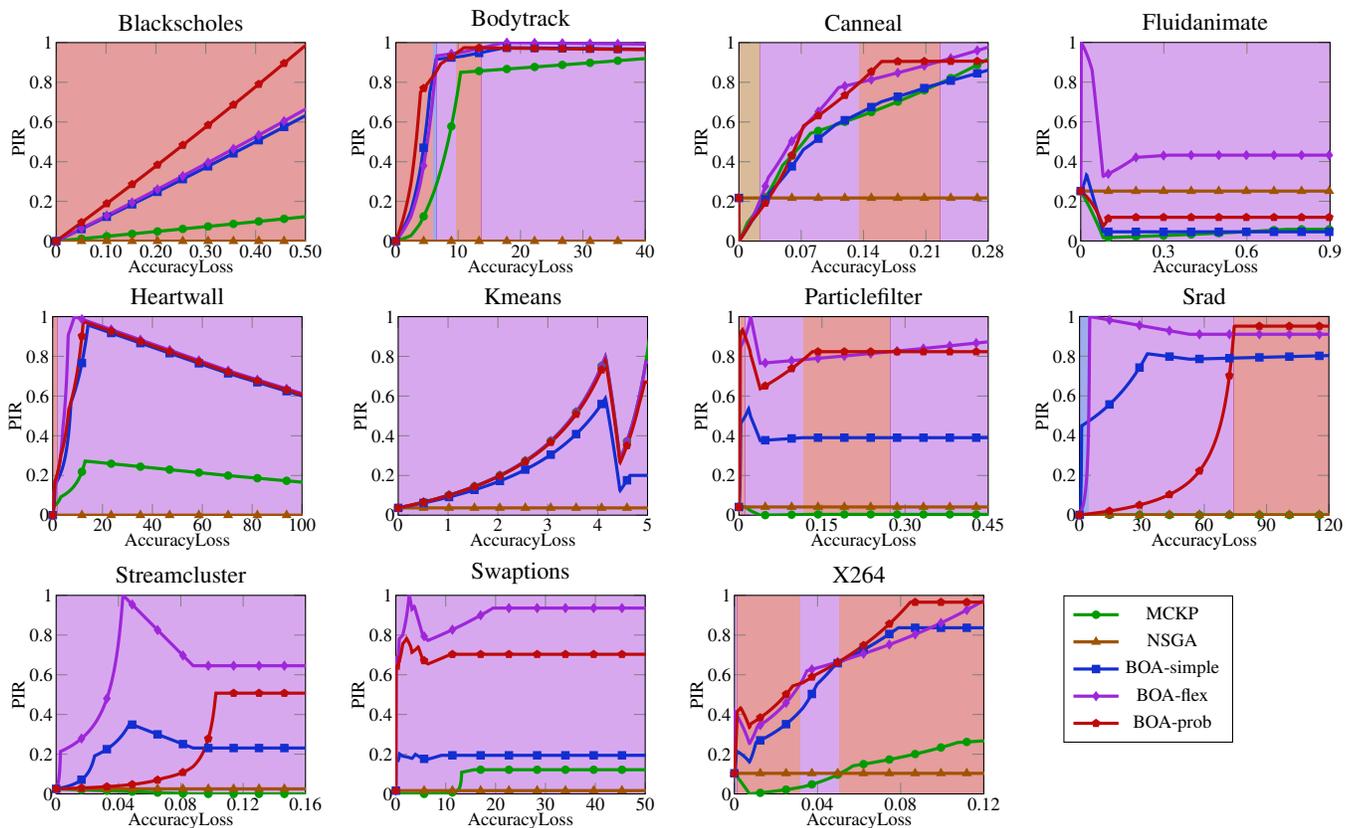

\newcolumntype{V}{>{\centering\arraybackslash} m{0.1\linewidth} }
\newcolumntype{C}[1]{>{\hsize=#1\hsize\centering\arraybackslash}X}%

\subfloat[(a) Pareto-optimal curves found by each exploration technique.]{
\hskip0.0cm \begin{tabular*}{\textwidth}{c @{\extracolsep{\fill}} c@{\extracolsep{\fill}} c@{\extracolsep{\fill}} c@{\extracolsep{\fill}}}
\subfloat{\input{img/cs2configSpace/Blackscholes-configSpace.tex}} &
\subfloat{\input{img/cs2configSpace/Bodytrack-configSpace.tex} } &
\subfloat{\hspace*{-85pt} \input{img/cs2configSpace/Canneal-configSpace.tex}} &
\subfloat{\hspace*{-85pt} \input{img/cs2configSpace/Fluidanimate-configSpace.tex}} \\[-3.5ex]

\subfloat{\input{img/cs2configSpace/Heartwall-configSpace.tex} } &
\subfloat{\input{img/cs2configSpace/Kmeans-configSpace.tex} } &
\subfloat{\hspace*{-85pt} \input{img/cs2configSpace/Particlefilter-configSpace.tex}} &
\subfloat{\hspace*{-85pt} \input{img/cs2configSpace/Srad-configSpace.tex} } \\[-3.5ex]

\subfloat{\input{img/cs2configSpace/Streamcluster-configSpace.tex}} &
\subfloat{\input{img/cs2configSpace/Swaptions-configSpace.tex} } &
\subfloat{\input{img/cs2configSpace/X264-configSpace.tex}} \\[-2.5ex]

\end{tabular*}
\label{fig:cs2configSpace}
}

\subfloat[(b) VIPER comparison of MCKP, NSGA-II, and different versions of \system{}.]{
\hskip0.0cm \begin{tabular*}{\textwidth}{c @{\extracolsep{\fill}} c@{\extracolsep{\fill}} c@{\extracolsep{\fill}} c@{\extracolsep{\fill}}}
\subfloat{\input{img/cs2viperLines/Blackscholes-cs2viperLine.tex}} &
\subfloat{\input{img/cs2viperLines/Bodytrack-cs2viperLine.tex} } &
\subfloat{\hspace*{-85pt} \input{img/cs2viperLines/Canneal-cs2viperLine.tex}} &
\subfloat{\hspace*{-85pt} \input{img/cs2viperLines/Fluidanimate-cs2viperLine.tex}} \\[-3.5ex]

\subfloat{\input{img/cs2viperLines/Heartwall-cs2viperLine.tex} } &
\subfloat{\input{img/cs2viperLines/Kmeans-cs2viperLine.tex} } &
\subfloat{\hspace*{-85pt} \input{img/cs2viperLines/Particlefilter-cs2viperLine.tex}} &
\subfloat{\hspace*{-85pt} \input{img/cs2viperLines/Srad-cs2viperLine.tex}} \\[-3.5ex]

\subfloat{\input{img/cs2viperLines/Streamcluster-cs2viperLine.tex}} &
\subfloat{\input{img/cs2viperLines/Swaptions-cs2viperLine.tex} } &
\subfloat{\input{img/cs2viperLines/X264-cs2viperLine.tex}} \\[-2.5ex]

\end{tabular*}
\label{fig:cs2viperLines}
}

\vspace{-2ex}

\caption{Comparison of MCKP, NSGA-II, and \system{} with different thresholds.  (a) shows
  comparison by Pareto-efficient frontiers and (b) shows comparison by
  VIPER.}
\label{fig:compare}
\end{figure*}

\figref{cs2configSpace} shows the Pareto-efficient points for each
benchmark and search method. The y-axis shows runtime (normalized to
the default configuration) and the x-axis represents accuracy loss. We 
use median runtime across test inputs.
These figures display the range of runtime and accuracy loss that a
method can achieve. For instance, NSGA-II and MCKP cannot provide
normalized runtime less than 78.1\% and 55.5\% of the default
configuration respectively for \texttt{particlefilter}.


Figure \textcolor{red}{5b} illustrates the VIPER comparison of  
NSGA-II, MCKP, and different variations of
\system{}. The y-axis shows performance improvement ratio while the
x-axis shows accuracy loss. We use NSGA-II as the baseline, so it is
represented by a horizontal line.  For the same accuracy loss, lines
above that horizontal represent better (more efficient)
configurations, and lines below represent configurations worse than
those found by NSGA-II. 
For most applications, MCKP stays below the horizontal, meaning it
is worse than  NSGA-II. By comparing
\system{}-simple and MCKP performance improvement ratio lines, we see
that \system{}-simple outperforms MCKP.

From the \cmpsystem{} plots we also find the maximum and minimum
performance improvement over NSGA-II. Considering
\texttt{fluidanimate}, the NSGA-II line is at 0.25 indicating that the
maximum performance is 4$\times$ better than NSGA-II, and minimum
performance is 25\% worse. In fact, for every benchmark \system{}-flex
finds at least one configuration with higher performance for the same accuracy.

Whenever NSGA-II locates more Pareto-efficient points than
\system{}-simple, by expanding the Pareto-efficient configurations we
reduce the performance improvement ratio gap. Benchmarks
\texttt{heartwall}, \texttt{kmeans}, and \texttt{particlefilter}
demonstrate how expanding the combined configurations provides higher
Pareto-efficiency. In total, we find by increasing the threshold,
lines of \system{}-flex are above the NSGA-II line more than 95\% of
the time. 
%
\emph{These results provide visual
  confirmation that \system{} not only finds a greater number of
  efficient points than prior techniques, but \system{}'s points are
  also significantly better, representing much more efficient
  trade-offs. Furthermore, we believe this case study provides 
  further evidence of VIPER's value, as the VIPER charts are 
  visually intuitive in Figure 5b, but the Pareto frontiers (Figure 5a) 
  do not immediately show which framework is best at a given accuracy 
  or by how much. }
  
\subsection{Exploration Time} 
\label{sec:exploredConfigsEvaluation}

\def\tabularxcolumn#1{m{#1}}
\newcolumntype{C}[1]{>{\hsize=#1\hsize\centering\small\arraybackslash}X}%
\newcolumntype{D}[1]{>{\hsize=#1\hsize\centering\bfseries\footnotesize\arraybackslash}X}%
\newcolumntype{E}[1]{>{\hsize=#1\hsize\centering\footnotesize\arraybackslash}X}%
\newcolumntype{F}[1]{>{\hsize=#1\hsize\centering\tiny\arraybackslash}X}%

\begin{table*}[t]
\centering
\caption{Number of Explored Configurations.}

  \resizebox{0.85\textwidth}{!}{
\begin{tabularx}{\textwidth}{|D{1.60}|E{0.85}|E{0.85}|E{0.85}|E{0.91}|E{0.91}|E{0.91}|E{1}|E{1}|E{0.91}|E{1.21}|}
\hline 
\cellcolor[gray]{0.8}{Benchmarks}  & \cellcolor[gray]{0.8} {LP} & \cellcolor[gray]{0.8} {PD} & \cellcolor[gray]{0.8} {AML} & \cellcolor[gray]{0.8} {MCKP} & \cellcolor[gray]{0.8} {NSGA-II} & \cellcolor[gray]{0.8} {\system{}-simple} & \cellcolor[gray]{0.8}  {\system{}-flex {\footnotesize Th=0.05}} & \cellcolor[gray]{0.8} {\system{}-flex {\footnotesize Th=0.1}} & \cellcolor[gray]{0.8} {\system{}-prob} &  Combined Configs \\ 
\hline 
Blackscholes & 20 & - & 216 & 5 & 240 & 3 & 12 & 18 & 27 & 4,320 \\ 
\hline 
Bodytrack & 768 & 200 & 36 & 15 & 800 & 30 & 224 & 256 & 144 &  5,529,600 \\ 
\hline 
Canneal & 7 & 525 & - & 23 & 480 & 33 & 102 & 165 & 110 &  3,675 \\ 
\hline 
Fluidanimate & 144 & 20 & 6 & 11 & 180 & 24  & 216 & 672 & 168 &  17,280 \\ 
\hline 
Heartwall & 256 & 320 & - & 19 & 400 & 98  & 665 & 722 & 777 &  81,920 \\ 
\hline 
Kmeans & 120 & 100 & - & 11 & 200 & 32 & 216 & 216 & 192 &  12,000 \\ 
\hline 
Particlefilter & 200 & 380 & 216 & 12 & 1000 & 240 & 1760 & 7200 & 1728 &  16,416,000 \\ 
\hline 
Srad & 256 & 10 & 36 & 11 & 320 & 48 & 288 & 396 & 160 &  92,160 \\ 
\hline 
Streamcluster & 384 & 256 & 6 & 9 & 480 & 80  & 392 & 1380 & 640 &  589,824 \\ 
\hline 
Swaptions & 768 & 100 & 36 & 15 & 1000 & 330  & 2310 & 11025 & 1584 &  2,764,800 \\ 
\hline 
X264 & 768 & 400 & - & 17 & 1000 & 45 & 345 & 400 & 405 &  307,200 \\ 
\hline 
\end{tabularx} 
}
\label{tbl:exploredConfigs}
\end{table*}

The Pareto-efficiency of the located points depends on exploration
time.  While Figures \ref{fig:convergence} and \ref{fig:compare} show
that \system{} produces better configurations than other
techniques, it is important to know if that gain comes
from exploring more points or from a better exploration strategy.
Table \ref{tbl:exploredConfigs} presents the number of configurations
explored for each benchmark for different methods, including different
thresholds for \system{}-flex. 
To get an estimate of the time spent exploring the combined 
trade-off space for a specific benchmark, we can multiply 
the number of explored configurations by the average runtime 
(from the last row in Table 1).
Comparing NSGA-II and \system{}-simple
across all benchmarks, NSGA-II explores 2.05\% of all possible
configurations, while \system{}-simple explores about 14$\times$ less.  
\system{}-flex and
\system{}-prob only search the 0.682\% and 0.692\% of all possible
configurations, respectively.  
 \emph{These results
  indicate that \system{} not only finds better combinations of
  approximate frameworks, it does so with less searching.}

Since MCKP only chooses configurations from individual Pareto-optimal
curves rather than merging the configurations, the number of explored
configurations stays very low. In the worst case, MCKP explores up to
the sum of the Pareto-optimal points of PowerDial, Loop Perforation,
and the Approximate Math Library.  Unfortunately, while MCKP searches
a small space, it is too small to find many useful points.

\subsection{Input Sensitivity}
\label{sec:inputSensitivity}
Since exhaustive exploration is not feasible, we use training and 
test data to ensure robustness of \system{} on unseen data. We show 
how well the behavior on
training inputs predicts that on test inputs.  For each search method,
we take the normalized runtime and accuracy loss, compute a linear
least squares fit of training data to test data, and compute the
correlation coefficient of each fit.  Higher correlation coefficients
imply greater robustness; \ie{} the behavior of configurations
found during training data is a good predictor of test behavior

Table \ref{tbl:correlationTableAccuLoss} shows the correlation
coefficient ($R$-values) for accuracy loss for each exploration method
per benchmark. Table \ref{tbl:correlationTableRuntime} shows the
$R$-values for runtime.  By harmonic mean, \system{} has higher
consistency of accuracy loss and normalized runtime comparing to
NSGA-II by 17\% and 64\% respectively.  Since MCKP evaluates few
configurations, its predictions are quite robust---one advantage 
of MCKP over other techniques.

While some
benchmarks such as \texttt{fluidanimate} and \texttt{streamcluster}
clearly stress the difference between training and test inputs.
NSGA-II's heuristic approach can select configurations
for the training data that produce bad results on the test
data. In contrast, \system{} not only finds more efficient
configurations, its results are also much more robust when applied to
new inputs, producing uniformly high $R$-values.
\emph{These results indicate that \system{} is a
sound method for combining approximation frameworks.}

\def\tabularxcolumn#1{m{#1}}

\newcolumntype{M}[1]{>{\centering\arraybackslash}m{#1}}

\begin{table}[t]
  \centering
  \caption{Correlation coefficients for accuracy loss.}
  \resizebox{0.40\textwidth}{!}{
\begin{tabularx}{0.45\textwidth}{|D{1.40}|E{0.88}|E{0.90}|E{1.02}|E{0.90}|E{0.90}|} 
\hline 
\cellcolor[gray]{0.7} {\tiny Benchmark} & \cellcolor[gray]{0.7} {\tiny MCKP} &  \cellcolor[gray]{0.7} {\tiny NSGA-II} & \cellcolor[gray]{0.7} {\tiny \system{}-simple} & \cellcolor[gray]{0.7} {\tiny \system{}-flex} & \cellcolor[gray]{0.7} {\tiny \system{}-prob} \\ 
 \hline 
{\scriptsize Blackscholes} & 1 & 1 & 1 & 1 & 1\\
\hline
{\scriptsize Bodytrack} & 0.992 & 0.972 & 0.990  & 0.989 & 0.949 \\
\hline
{\scriptsize Canneal} & 0.999 & 0.999 & 1 & 1 & 1 \\ 
\hline
{\scriptsize Fluidanimate} & 0.539 & 0.463 & 0.594  & 0.943 & 0.772 \\
\hline
{\scriptsize Heartwall} & 0.951 & 0.341 & 0.964  & 0.999 & 0.959 \\
\hline
{\scriptsize Kmeans} & 0.999 & 0.996 & 0.999  & 1 & 0.999 \\
\hline
{\scriptsize Particlefilter} & 0.933 & 0.999 & 0.822  & 0.997 & 0.997 \\
\hline
{\scriptsize Srad} & 1 & 1 & 0.958 &  0.927 & 0.999 \\
\hline
{\scriptsize Streamcluster} & 1 & 0.403 & 0.573 & 0.873 & 0.562 \\
\hline
{\scriptsize Swaptions} & 1 & 0.999 & 0.999 & 0.999 & 0.999 \\
\hline
{\scriptsize X264} & 0.888 & 0.841 & 0.992 & 0.938  & 0.983\\
\hline
{\scriptsize Average} & 0.908 & 0.696 & 0.863 & 0.968 & 0.928 \\
\hline 
\end{tabularx} 
}
\label{tbl:correlationTableAccuLoss}
\end{table}

\def\tabularxcolumn#1{m{#1}}

\newcolumntype{M}[1]{>{\centering\arraybackslash}m{#1}}

\begin{table}[t]
  \centering
  \caption{Correlation coefficients for normalized runtime.}
  \resizebox{0.40\textwidth}{!}{
\begin{tabularx}{0.45\textwidth}{|D{1.40}|E{0.88}|E{0.90}|E{1.02}|E{0.90}|E{0.90}|} 
\hline 
\cellcolor[gray]{0.7} {\tiny Benchmark}    & \cellcolor[gray]{0.7} {\tiny MCKP} &  \cellcolor[gray]{0.7} {\tiny NSGA-II} & \cellcolor[gray]{0.7} {\tiny \system{}-simple} & \cellcolor[gray]{0.7} {\tiny \system{}-flex} & \cellcolor[gray]{0.7} {\tiny \system{}-prob} \\ 
  \hline
{\scriptsize  Blackscholes} & 0.993 & 0.725 & 1 & 0.988 & 0.982 \\
\hline
{\scriptsize Bodytrack} & 0.999 & 0.999 & 0.999 &  0.999 & 0.999 \\
\hline
{\scriptsize Canneal} & 0.985 & 0.319 & 0.988 &  0.989 & 0.998 \\
\hline 
{\scriptsize Fluidanimate} & 0.999 & 0.999 & 0.999 &  0.953 & 0.985 \\
\hline
{\scriptsize Heartwall} & 0.999 & 0.999 & 0.999 &  0.999 & 0.999 \\
\hline
{\scriptsize Kmeans} & 0.981 & 0.935 & 0.992 &  1 & 0.997 \\
\hline
{\scriptsize Particlefilter} & 0.992 & 0.987 & 0.998 &  0.999 & 0.998 \\
\hline
{\scriptsize Srad} & 0.998 & 0.746 & 0.999  & 0.999 & 0.999 \\
\hline
{\scriptsize Streamcluster} & 1 & 0.056 & 0.998 &  0.999 & 0.999 \\
\hline
{\scriptsize Swaptions} & 1 & 0.991 & 0.999 & 0.973 & 0.971 \\
\hline
{\scriptsize X264} & 0.998 & 0.995 & 0.983 & 0.980 & 0.998 \\
\hline
{\scriptsize Average } & 0.995 & 0.358 & 0.996 & 0.989 & 0.993 \\
 \hline 
 \end{tabularx} 
 }
\label{tbl:correlationTableRuntime}
\end{table}

\captionsetup[subfigure]{labelformat=empty}

\subsection{Combination Distribution}

\begin{table}[t]
  \centering
\caption{Combinations of approximation frameworks found by BOA.}
\resizebox{0.35\textwidth}{!}{
\begin{tabularx}{0.40\textwidth}{|D{1.32}|E{0.92}|E{0.92}|E{0.92}|E{0.92}|} 
\hline 
\cellcolor[gray]{0.7} {\tiny Benchmark} & \cellcolor[gray]{0.7} {\tiny LP (Pareto-opt)} &  \cellcolor[gray]{0.7} {\tiny PD (Pareto-opt)} & \cellcolor[gray]{0.7} {\tiny AML (Pareto-opt)} & \cellcolor[gray]{0.7} {\tiny \system{}-simple} \\ 
  \hline
{\scriptsize  Blackscholes} & 1 & - & 3 & 3 \\
\hline
{\scriptsize Bodytrack} & 6 & 5 & 1 & 30 \\
\hline
{\scriptsize Canneal} & 3 & 11 & - & 33 \\
\hline 
{\scriptsize Fluidanimate} & 4 & 3 & 2 & 24 \\
\hline
{\scriptsize Heartwall} & 7 & 14 & - & 98 \\
\hline
{\scriptsize Kmeans} & 4 & 8 & - & 32 \\
\hline
{\scriptsize Particlefilter} & 6 & 8 & 5 & 240 \\
\hline
{\scriptsize Srad} & 2 & 8 & 3 & 48 \\
\hline
{\scriptsize Streamcluster} & 8 & 5 & 2 & 80 \\
\hline
{\scriptsize Swaptions} & 11 & 5 & 6 & 330 \\
\hline
{\scriptsize X264} & 9 & 5 & - & 45 \\
 \hline 
 \end{tabularx} 
 }
\label{tbl:distributedConfigs}
\end{table}

When \system{} combines frameworks, it considers
multiple configurations from each rather than choosing from
one or two only. Table \ref{tbl:distributedConfigs}
includes the number of configurations \system{}-simple selects from
each framework to generate the new, combined trade-off
space. As mentioned in \secref{compareAFPIR}, the
Approximate Math Library is never better than Loop Perforation
or PowerDial in any range of accuracy loss. However, BOA uses the
Approximate Math Library in combination with Loop Perforation and
PowerDial for 7 out of the 11 applications.  
\emph{These
  results show that there is real benefit to combining
  frameworks, as even the Approximate Math Library---which is
  uniformly the worst of the three techniques by
  themselves---contributes to Pareto-efficient points in the combined
  space found by \system{}.}

\section{Conclusion}
\label{sec:conclusion}

A proliferation of approximation
frameworks have recently appeared that exploit different configurable
parameters to trade reduced accuracy for decreased resource consumption.
This paper proposes methods for both comparing and combining different
frameworks.  \cmpsystem{} is a visualization tool for comparing
approximation frameworks across their entire range of available
accuracies.  We show this tool is useful for comparing existing
approximation frameworks regardless of their type and applied system
level.  \system{} is a family of algorithms that combine approximation 
frameworks
and quickly locate Pareto-efficient
configuration combination. 


\paragraph*{Acknowledgments}
The effort on this project is funded by the U.S.  Government under the
DARPA BRASS program and by a DOE Early Career Award. Additional
funding comes from the NSF (CCF-1439156, CNS-1526304, CCF-1823032,
CNS-1764039.


\newpage
\bibliographystyle{IEEEtran} 
\bibliography{IEEEabrv,bibliography}

\begin{thebibliography}{10}
\providecommand{\url}[1]{#1}
\csname url@samestyle\endcsname
\providecommand{\newblock}{\relax}
\providecommand{\bibinfo}[2]{#2}
\providecommand{\BIBentrySTDinterwordspacing}{\spaceskip=0pt\relax}
\providecommand{\BIBentryALTinterwordstretchfactor}{4}
\providecommand{\BIBentryALTinterwordspacing}{\spaceskip=\fontdimen2\font plus
\BIBentryALTinterwordstretchfactor\fontdimen3\font minus
  \fontdimen4\font\relax}
\providecommand{\BIBforeignlanguage}[2]{{%
\expandafter\ifx\csname l@#1\endcsname\relax
\typeout{** WARNING: IEEEtran.bst: No hyphenation pattern has been}%
\typeout{** loaded for the language `#1'. Using the pattern for}%
\typeout{** the default language instead.}%
\else
\language=\csname l@#1\endcsname
\fi
#2}}
\providecommand{\BIBdecl}{\relax}
\BIBdecl

\bibitem{Palem2003}
\BIBentryALTinterwordspacing
K.~V. Palem, ``Energy aware algorithm design via probabilistic computing: From
  algorithms and models to moore's law and novel (semiconductor) devices,'' in
  \emph{Proceedings of the 2003 International Conference on Compilers,
  Architecture and Synthesis for Embedded Systems}, ser. CASES '03.\hskip 1em
  plus 0.5em minus 0.4em\relax New York, NY, USA: ACM, 2003, pp. 113--116.
  [Online]. Available: \url{http://doi.acm.org/10.1145/951710.951712}
\BIBentrySTDinterwordspacing

\bibitem{lingamneni2013}
A.~Lingamneni, C.~Enz, K.~Palem, and C.~Piguet, ``Designing energy-efficient
  arithmetic operators using inexact computing,'' \emph{Journal of Low Power
  Electronics}, vol.~9, no.~1, pp. 141--153, 2013.

\bibitem{Ingole2015}
A.~Ingole, B.~Maiti, J.~Augustine, and K.~Palem, ``Does customizing inexactness
  help over simplistic precision (bit-width) reduction? a case study,'' in
  \emph{Compilers, Architecture and Synthesis for Embedded Systems (CASES),
  2015 International Conference on}, Oct 2015, pp. 33--34.

\bibitem{Chippa2013}
V.~K. Chippa, S.~Venkataramani, S.~T. Chakradhar, K.~Roy, and A.~Raghunathan,
  ``Approximate computing: An integrated hardware approach,'' in \emph{2013
  Asilomar Conference on Signals, Systems and Computers}, Nov 2013, pp.
  111--117.

\bibitem{Muralidharan2016}
\BIBentryALTinterwordspacing
S.~Muralidharan, A.~Roy, M.~Hall, M.~Garland, and P.~Rai,
  ``Architecture-adaptive code variant tuning,'' \emph{SIGPLAN Not.}, vol.~51,
  no.~4, pp. 325--338, Mar. 2016. [Online]. Available:
  \url{http://doi.acm.org/10.1145/2954679.2872411}
\BIBentrySTDinterwordspacing

\bibitem{Chakrapani2006}
L.~N. Chakrapani, B.~E.~S. Akgul, S.~Cheemalavagu, P.~Korkmaz, K.~V. Palem, and
  B.~Seshasayee, ``Ultra-efficient (embedded) soc architectures based on
  probabilistic cmos (pcmos) technology,'' in \emph{Proceedings of the Design
  Automation Test in Europe Conference}, vol.~1, March 2006, pp. 1--6.

\bibitem{samadi2014}
M.~Samadi, D.~A. Jamshidi, J.~Lee, and S.~Mahlke, ``Paraprox: Pattern-based
  approximation for data parallel applications,'' in \emph{ACM SIGARCH Computer
  Architecture News}, vol.~42, no.~1.\hskip 1em plus 0.5em minus 0.4em\relax
  ACM, 2014, pp. 35--50.

\bibitem{Esmaeilzadeh2012}
\BIBentryALTinterwordspacing
H.~Esmaeilzadeh, A.~Sampson, L.~Ceze, and D.~Burger, ``Neural acceleration for
  general-purpose approximate programs,'' in \emph{Proceedings of the 2012 45th
  Annual IEEE/ACM International Symposium on Microarchitecture}, ser.
  MICRO-45.\hskip 1em plus 0.5em minus 0.4em\relax Washington, DC, USA: IEEE
  Computer Society, 2012, pp. 449--460. [Online]. Available:
  \url{http://dx.doi.org/10.1109/MICRO.2012.48}
\BIBentrySTDinterwordspacing

\bibitem{Temam2012}
\BIBentryALTinterwordspacing
O.~Temam, ``A defect-tolerant accelerator for emerging high-performance
  applications,'' in \emph{Proceedings of the 39th Annual International
  Symposium on Computer Architecture}, ser. ISCA '12.\hskip 1em plus 0.5em
  minus 0.4em\relax Washington, DC, USA: IEEE Computer Society, 2012, pp.
  356--367. [Online]. Available:
  \url{http://dl.acm.org/citation.cfm?id=2337159.2337200}
\BIBentrySTDinterwordspacing

\bibitem{Esmaeilzadeh2006}
H.~Esmaeilzadeh, P.~Saeedi, B.~N. Araabi, C.~Lucas, and S.~M. Fakhraie,
  ``Neural network stream processing core (nnsp) for embedded systems,'' in
  \emph{2006 IEEE International Symposium on Circuits and Systems}, May 2006,
  pp. 4 pp.--2776.

\bibitem{bornholt2014}
J.~Bornholt, T.~Mytkowicz, and K.~S. McKinley, ``Uncertain< t>: A first-order
  type for uncertain data,'' \emph{ACM SIGPLAN Notices}, vol.~49, no.~4, pp.
  51--66, 2014.

\bibitem{Kansal2013}
\BIBentryALTinterwordspacing
A.~Kansal, S.~Saponas, A.~B. Brush, K.~S. McKinley, T.~Mytkowicz, and R.~Ziola,
  ``The latency, accuracy, and battery (lab) abstraction: Programmer
  productivity and energy efficiency for continuous mobile context sensing,''
  in \emph{Proceedings of the 2013 ACM SIGPLAN International Conference on
  Object Oriented Programming Systems Languages \&\#38; Applications}, ser.
  OOPSLA '13.\hskip 1em plus 0.5em minus 0.4em\relax New York, NY, USA: ACM,
  2013, pp. 661--676. [Online]. Available:
  \url{http://doi.acm.org/10.1145/2509136.2509541}
\BIBentrySTDinterwordspacing

\bibitem{Sampson2011}
\BIBentryALTinterwordspacing
A.~Sampson, W.~Dietl, E.~Fortuna, D.~Gnanapragasam, L.~Ceze, and D.~Grossman,
  ``Enerj: Approximate data types for safe and general low-power computation,''
  in \emph{Proceedings of the 32Nd ACM SIGPLAN Conference on Programming
  Language Design and Implementation}, ser. PLDI '11.\hskip 1em plus 0.5em
  minus 0.4em\relax New York, NY, USA: ACM, 2011, pp. 164--174. [Online].
  Available: \url{http://doi.acm.org/10.1145/1993498.1993518}
\BIBentrySTDinterwordspacing

\bibitem{Oh2013}
\BIBentryALTinterwordspacing
T.~Oh, H.~Kim, N.~P. Johnson, J.~W. Lee, and D.~I. August, ``Practical
  automatic loop specialization,'' \emph{SIGPLAN Not.}, vol.~48, no.~4, pp.
  419--430, Mar. 2013. [Online]. Available:
  \url{http://doi.acm.org/10.1145/2499368.2451161}
\BIBentrySTDinterwordspacing

\bibitem{ansel2011}
J.~Ansel, Y.~L. Wong, C.~Chan, M.~Olszewski, A.~Edelman, and S.~Amarasinghe,
  ``Language and compiler support for auto-tuning variable-accuracy
  algorithms,'' in \emph{Proceedings of the 9th Annual IEEE/ACM International
  Symposium on Code Generation and Optimization}.\hskip 1em plus 0.5em minus
  0.4em\relax IEEE Computer Society, 2011, pp. 85--96.

\bibitem{Baek2010}
\BIBentryALTinterwordspacing
W.~Baek and T.~M. Chilimbi, ``Green: A framework for supporting
  energy-conscious programming using controlled approximation,'' \emph{SIGPLAN
  Not.}, vol.~45, no.~6, pp. 198--209, Jun. 2010. [Online]. Available:
  \url{http://doi.acm.org/10.1145/1809028.1806620}
\BIBentrySTDinterwordspacing

\bibitem{Fousse2007}
\BIBentryALTinterwordspacing
L.~Fousse, G.~Hanrot, V.~Lef\`{e}vre, P.~P{\'e}lissier, and P.~Zimmermann,
  ``Mpfr: A multiple-precision binary floating-point library with correct
  rounding,'' \emph{ACM Trans. Math. Softw.}, vol.~33, no.~2, Jun. 2007.
  [Online]. Available: \url{http://doi.acm.org/10.1145/1236463.1236468}
\BIBentrySTDinterwordspacing

\bibitem{Kwon2009}
\BIBentryALTinterwordspacing
T.-J. Kwon and J.~Draper, ``Floating-point division and square root using a
  taylor-series expansion algorithm,'' \emph{Microelectronics Journal},
  vol.~40, no.~11, pp. 1601 -- 1605, 2009, international Conference on
  MicroelectronicsDigital and Mixed-Signal Circuits and Systems. [Online].
  Available:
  \url{http://www.sciencedirect.com/science/article/pii/S0026269209000500}
\BIBentrySTDinterwordspacing

\bibitem{Zhou2011}
\BIBentryALTinterwordspacing
A.~Zhou, B.-Y. Qu, H.~Li, S.-Z. Zhao, P.~N. Suganthan, and Q.~Zhang,
  ``Multiobjective evolutionary algorithms: A survey of the state of the art,''
  \emph{Swarm and Evolutionary Computation}, vol.~1, no.~1, pp. 32 -- 49, 2011.
  [Online]. Available:
  \url{http://www.sciencedirect.com/science/article/pii/S2210650211000058}
\BIBentrySTDinterwordspacing

\bibitem{Holzer2007}
M.~Holzer, B.~Knerr, and M.~Rupp, ``Design space exploration with evolutionary
  multi-objective optimisation,'' in \emph{2007 International Symposium on
  Industrial Embedded Systems}, July 2007, pp. 126--133.

\bibitem{Ascia2004}
G.~Ascia, V.~Catania, and M.~Palesi, ``A ga-based design space exploration
  framework for parameterized system-on-a-chip platforms,'' \emph{IEEE
  Transactions on Evolutionary Computation}, vol.~8, no.~4, pp. 329--346, Aug
  2004.

\bibitem{Marti2007}
\BIBentryALTinterwordspacing
L.~Mart\'{\i}, J.~Garc\'{\i}a, A.~Berlanga, and J.~M. Molina, ``A cumulative
  evidential stopping criterion for multiobjective optimization evolutionary
  algorithms,'' in \emph{Proceedings of the 9th Annual Conference Companion on
  Genetic and Evolutionary Computation}, ser. GECCO '07.\hskip 1em plus 0.5em
  minus 0.4em\relax New York, NY, USA: ACM, 2007, pp. 2835--2842. [Online].
  Available: \url{http://doi.acm.org/10.1145/1274000.1274053}
\BIBentrySTDinterwordspacing

\bibitem{Marti2009}
L.~Marti, J.~Garcia, A.~Berlanga, and J.~M. Molina, ``An approach to stopping
  criteria for multi-objective optimization evolutionary algorithms: The mgbm
  criterion,'' in \emph{2009 IEEE Congress on Evolutionary Computation}, May
  2009, pp. 1263--1270.

\bibitem{Sidiroglou-Douskos2011}
\BIBentryALTinterwordspacing
S.~Sidiroglou-Douskos, S.~Misailovic, H.~Hoffmann, and M.~Rinard, ``Managing
  performance vs. accuracy trade-offs with loop perforation,'' in
  \emph{Proceedings of the 19th ACM SIGSOFT Symposium and the 13th European
  Conference on Foundations of Software Engineering}, ser. ESEC/FSE '11.\hskip
  1em plus 0.5em minus 0.4em\relax New York, NY, USA: ACM, 2011, pp. 124--134.
  [Online]. Available: \url{http://doi.acm.org/10.1145/2025113.2025133}
\BIBentrySTDinterwordspacing

\bibitem{Hoffmann2011}
\BIBentryALTinterwordspacing
H.~Hoffmann, S.~Sidiroglou, M.~Carbin, S.~Misailovic, A.~Agarwal, and
  M.~Rinard, ``Dynamic knobs for responsive power-aware computing,'' in
  \emph{Proceedings of the Sixteenth International Conference on Architectural
  Support for Programming Languages and Operating Systems}, ser. ASPLOS
  XVI.\hskip 1em plus 0.5em minus 0.4em\relax New York, NY, USA: ACM, 2011, pp.
  199--212. [Online]. Available:
  \url{http://doi.acm.org/10.1145/1950365.1950390}
\BIBentrySTDinterwordspacing

\bibitem{Yang2003}
P.~Yang and F.~Catthoor, ``Pareto-optimization-based run-time task scheduling
  for embedded systems,'' in \emph{Hardware/Software Codesign and System
  Synthesis, 2003. First IEEE/ACM/IFIP International Conference on}, Oct 2003,
  pp. 120--125.

\bibitem{Deb2002}
K.~Deb, A.~Pratap, S.~Agarwal, and T.~Meyarivan, ``A fast and elitist
  multiobjective genetic algorithm: Nsga-ii,'' \emph{IEEE Transactions on
  Evolutionary Computation}, vol.~6, no.~2, pp. 182--197, Apr 2002.

\bibitem{Esmaeilzadeh2011}
\BIBentryALTinterwordspacing
H.~Esmaeilzadeh, A.~Sampson, L.~Ceze, and D.~Burger, ``Architecture support for
  disciplined approximate programming,'' \emph{SIGPLAN Not.}, vol.~47, no.~4,
  pp. 301--312, Mar. 2012. [Online]. Available:
  \url{http://doi.acm.org/10.1145/2248487.2151008}
\BIBentrySTDinterwordspacing

\bibitem{Omer}
\BIBentryALTinterwordspacing
Q.~Shi, H.~Hoffmann, and O.~Khan, ``A cross-layer multicore architecture to
  tradeoff program accuracy and resilience overheads,'' \emph{{IEEE} Comput.
  Archit. Lett.}, vol.~14, no.~2, pp. 85--89, 2015. [Online]. Available:
  \url{https://doi.org/10.1109/LCA.2014.2365204}
\BIBentrySTDinterwordspacing

\bibitem{Patterns2010}
\BIBentryALTinterwordspacing
M.~Rinard, H.~Hoffmann, S.~Misailovic, and S.~Sidiroglou, ``Patterns and
  statistical analysis for understanding reduced resource computing,'' in
  \emph{Proceedings of the ACM International Conference on Object Oriented
  Programming Systems Languages and Applications}, ser. OOPSLA '10.\hskip 1em
  plus 0.5em minus 0.4em\relax New York, NY, USA: Association for Computing
  Machinery, 2010, p. 806821. [Online]. Available:
  \url{https://doi.org/10.1145/1869459.1869525}
\BIBentrySTDinterwordspacing

\bibitem{ICSE2010}
\BIBentryALTinterwordspacing
S.~Misailovic, S.~Sidiroglou, H.~Hoffmann, and M.~Rinard, \emph{Quality of
  Service Profiling}.\hskip 1em plus 0.5em minus 0.4em\relax New York, NY, USA:
  Association for Computing Machinery, 2010, p. 2534. [Online]. Available:
  \url{https://doi.org/10.1145/1806799.1806808}
\BIBentrySTDinterwordspacing

\bibitem{Hoffmann2009}
H.~Hoffmann, S.~Misailovic, S.~Sidiroglou, A.~Agarwal, and M.~Rinard, ``Using
  code perforation to improve performance, reduce energy consumption, and
  respond to failures,'' no. MIT-CSAIL-TR-2009-042, 09 2009.

\bibitem{Samadi2013}
\BIBentryALTinterwordspacing
M.~Samadi, J.~Lee, D.~A. Jamshidi, A.~Hormati, and S.~Mahlke, ``Sage:
  Self-tuning approximation for graphics engines,'' in \emph{Proceedings of the
  46th Annual IEEE/ACM International Symposium on Microarchitecture}, ser.
  MICRO-46.\hskip 1em plus 0.5em minus 0.4em\relax New York, NY, USA: ACM,
  2013, pp. 13--24. [Online]. Available:
  \url{http://doi.acm.org/10.1145/2540708.2540711}
\BIBentrySTDinterwordspacing

\bibitem{Park2016}
\BIBentryALTinterwordspacing
J.~Park, E.~Amaro, D.~Mahajan, B.~Thwaites, and H.~Esmaeilzadeh, ``Axgames:
  Towards crowdsourcing quality target determination in approximate
  computing,'' \emph{SIGPLAN Not.}, vol.~51, no.~4, pp. 623--636, Mar. 2016.
  [Online]. Available: \url{http://doi.acm.org/10.1145/2954679.2872376}
\BIBentrySTDinterwordspacing

\bibitem{Mathar2009}
R.~J. Mathar, ``A java math. bigdecimal implementation of core mathematical
  functions,'' \emph{arXiv preprint arXiv:0908.3030}, 2009.

\bibitem{Abad2015}
\BIBentryALTinterwordspacing
A.~Abad, R.~Barrio, M.~Marco-Buzunariz, and M.~Rodríguez, ``Automatic
  implementation of the numerical taylor series method: A mathematica and sage
  approach,'' \emph{Applied Mathematics and Computation}, vol. 268, pp. 227 --
  245, 2015. [Online]. Available:
  \url{http://www.sciencedirect.com/science/article/pii/S0096300315008231}
\BIBentrySTDinterwordspacing

\bibitem{JouleGuard}
H.~Hoffmann, ``Jouleguard: Energy guarantees for approximate applications,'' in
  \emph{SOSP}, 2015.

\bibitem{Canino2018}
\BIBentryALTinterwordspacing
A.~Canino, Y.~D. Liu, and H.~Masuhara, ``Stochastic energy optimization for
  mobile gps applications,'' in \emph{Proceedings of the 2018 26th ACM Joint
  Meeting on European Software Engineering Conference and Symposium on the
  Foundations of Software Engineering}, ser. ESEC/FSE 2018.\hskip 1em plus
  0.5em minus 0.4em\relax New York, NY, USA: ACM, 2018, pp. 703--713. [Online].
  Available: \url{http://doi.acm.org/10.1145/3236024.3236076}
\BIBentrySTDinterwordspacing

\bibitem{Sui2016}
\BIBentryALTinterwordspacing
X.~Sui, A.~Lenharth, D.~S. Fussell, and K.~Pingali, ``Proactive control of
  approximate programs,'' \emph{SIGOPS Oper. Syst. Rev.}, vol.~50, no.~2, pp.
  607--621, Mar. 2016. [Online]. Available:
  \url{http://doi.acm.org/10.1145/2954680.2872402}
\BIBentrySTDinterwordspacing

\bibitem{ALERT1}
\BIBentryALTinterwordspacing
C.~Wan, H.~Hoffmann, S.~Lu, and M.~Maire, ``Orthogonalized {SGD} and nested
  architectures for anytime neural networks,'' in \emph{Proceedings of the 37th
  International Conference on Machine Learning}, ser. Proceedings of Machine
  Learning Research, H.~D. III and A.~Singh, Eds., vol. 119.\hskip 1em plus
  0.5em minus 0.4em\relax PMLR, 13--18 Jul 2020, pp. 9807--9817. [Online].
  Available: \url{http://proceedings.mlr.press/v119/wan20a.html}
\BIBentrySTDinterwordspacing

\bibitem{ALERT2}
\BIBentryALTinterwordspacing
C.~Wan, M.~Santriaji, E.~Rogers, H.~Hoffmann, M.~Maire, and S.~Lu, ``{ALERT}:
  Accurate learning for energy and timeliness,'' in \emph{2020 {USENIX} Annual
  Technical Conference ({USENIX} {ATC} 20)}.\hskip 1em plus 0.5em minus
  0.4em\relax {USENIX} Association, Jul. 2020, pp. 353--369. [Online].
  Available: \url{https://www.usenix.org/conference/atc20/presentation/wan}
\BIBentrySTDinterwordspacing

\bibitem{Sorber2007}
J.~Sorber, A.~Kostadinov, M.~Garber, M.~Brennan, M.~D. Corner, and E.~D.
  Berger, ``Eon: A language and runtime system for perpetual systems,'' in
  \emph{Proceedings of the 5th International Conference on Embedded Networked
  Sensor Systems}, ser. SenSys '07, 2007.

\bibitem{Proteus}
\BIBentryALTinterwordspacing
S.~Barati, F.~A. Bartha, S.~Biswas, R.~Cartwright, A.~Duracz, D.~S. Fussell,
  H.~Hoffmann, C.~Imes, J.~E. Miller, N.~Mishra, Arvind, D.~Nguyen, K.~V.
  Palem, Y.~Pei, K.~Pingali, R.~Sai, A.~Wright, Y.~Yang, and S.~Zhang,
  ``Proteus: Language and runtime support for self-adaptive software
  development,'' \emph{{IEEE} Software}, vol.~36, no.~2, pp. 73--82, 2019.
  [Online]. Available: \url{https://doi.org/10.1109/MS.2018.2884864}
\BIBentrySTDinterwordspacing

\bibitem{LAB}
A.~Kansal, S.~Saponas, A.~Brush, K.~S. McKinley, T.~Mytkowicz, and R.~Ziola,
  ``The latency, accuracy, and battery (lab) abstraction: programmer
  productivity and energy efficiency for continuous mobile context sensing,''
  in \emph{ACM SIGPLAN Notices}, 2013.

\bibitem{Ent}
\BIBentryALTinterwordspacing
A.~Canino and Y.~D. Liu, ``Proactive and adaptive energy-aware programming with
  mixed typechecking,'' in \emph{Proceedings of the 38th ACM SIGPLAN Conference
  on Programming Language Design and Implementation}, ser. PLDI 2017.\hskip 1em
  plus 0.5em minus 0.4em\relax New York, NY, USA: ACM, 2017, pp. 217--232.
  [Online]. Available: \url{http://doi.acm.org/10.1145/3062341.3062356}
\BIBentrySTDinterwordspacing

\bibitem{Ringenburg2015}
\BIBentryALTinterwordspacing
M.~Ringenburg, A.~Sampson, I.~Ackerman, L.~Ceze, and D.~Grossman, ``Monitoring
  and debugging the quality of results in approximate programs,'' \emph{SIGPLAN
  Not.}, vol.~50, no.~4, pp. 399--411, Mar. 2015. [Online]. Available:
  \url{http://doi.acm.org/10.1145/2775054.2694365}
\BIBentrySTDinterwordspacing

\bibitem{Carbin2013}
\BIBentryALTinterwordspacing
M.~Carbin, S.~Misailovic, and M.~C. Rinard, ``Verifying quantitative
  reliability for programs that execute on unreliable hardware,'' in
  \emph{Proceedings of the 2013 ACM SIGPLAN International Conference on Object
  Oriented Programming Systems Languages \&\#38; Applications}, ser. OOPSLA
  '13.\hskip 1em plus 0.5em minus 0.4em\relax New York, NY, USA: ACM, 2013, pp.
  33--52. [Online]. Available: \url{http://doi.acm.org/10.1145/2509136.2509546}
\BIBentrySTDinterwordspacing

\bibitem{Darulova2013}
\BIBentryALTinterwordspacing
E.~Darulova, V.~Kuncak, R.~Majumdar, and I.~Saha, ``Synthesis of fixed-point
  programs,'' in \emph{Proceedings of the Eleventh ACM International Conference
  on Embedded Software}, ser. EMSOFT '13.\hskip 1em plus 0.5em minus
  0.4em\relax Piscataway, NJ, USA: IEEE Press, 2013, pp. 22:1--22:10. [Online].
  Available: \url{http://dl.acm.org/citation.cfm?id=2555754.2555776}
\BIBentrySTDinterwordspacing

\bibitem{CoAdapt}
H.~Hoffmann, ``Coadapt: Predictable behavior for accuracy-aware applications
  running on power-aware systems,'' in \emph{26th Euromicro Conference on
  Real-Time Systems, {ECRTS} 2014, Madrid, Spain, July 8-11, 2014}, 2014, pp.
  223--232.

\bibitem{FSE2017}
\BIBentryALTinterwordspacing
M.~Maggio, A.~V. Papadopoulos, A.~Filieri, and H.~Hoffmann, ``Automated control
  of multiple software goals using multiple actuators,'' in \emph{Proceedings
  of the 2017 11th Joint Meeting on Foundations of Software Engineering,
  {ESEC/FSE} 2017, Paderborn, Germany, September 4-8, 2017}, 2017, pp.
  373--384. [Online]. Available: \url{https://doi.org/10.1145/3106237.3106247}
\BIBentrySTDinterwordspacing

\bibitem{FSE2015}
\BIBentryALTinterwordspacing
A.~Filieri, H.~Hoffmann, and M.~Maggio, ``Automated multi-objective control for
  self-adaptive software design,'' in \emph{Proceedings of the 2015 10th Joint
  Meeting on Foundations of Software Engineering, {ESEC/FSE} 2015, Bergamo,
  Italy, August 30 - September 4, 2015}, E.~D. Nitto, M.~Harman, and
  P.~Heymans, Eds.\hskip 1em plus 0.5em minus 0.4em\relax {ACM}, 2015, pp.
  13--24. [Online]. Available: \url{https://doi.org/10.1145/2786805.2786833}
\BIBentrySTDinterwordspacing

\bibitem{Meantime}
A.~Farrell and H.~Hoffmann, ``{MEANTIME:} achieving both minimal energy and
  timeliness with approximate computing,'' in \emph{2016 {USENIX} Annual
  Technical Conference, {USENIX} {ATC} 2016, Denver, CO, USA, June 22-24,
  2016.}, 2016, pp. 421--435.

\bibitem{TAAS2017}
\BIBentryALTinterwordspacing
A.~Filieri, M.~Maggio, K.~Angelopoulos, N.~D'Ippolito, I.~Gerostathopoulos,
  A.~B. Hempel, H.~Hoffmann, P.~Jamshidi, E.~Kalyvianaki, C.~Klein, F.~Krikava,
  S.~Misailovic, A.~V. Papadopoulos, S.~Ray, A.~M. Sharifloo, S.~Shevtsov,
  M.~Ujma, and T.~Vogel, ``Control strategies for self-adaptive software
  systems,'' \emph{{ACM} Trans. Auton. Adapt. Syst.}, vol.~11, no.~4, pp.
  24:1--24:31, 2017. [Online]. Available: \url{https://doi.org/10.1145/3024188}
\BIBentrySTDinterwordspacing

\bibitem{ASPLOS2018}
\BIBentryALTinterwordspacing
S.~Wang, C.~Li, H.~Hoffmann, S.~Lu, W.~Sentosa, and A.~I. Kistijantoro,
  ``Understanding and auto-adjusting performance-sensitive configurations,'' in
  \emph{Proceedings of the Twenty-Third International Conference on
  Architectural Support for Programming Languages and Operating Systems,
  {ASPLOS} 2018, Williamsburg, VA, USA, March 24-28, 2018}, X.~Shen, J.~Tuck,
  R.~Bianchini, and V.~Sarkar, Eds.\hskip 1em plus 0.5em minus 0.4em\relax
  {ACM}, 2018, pp. 154--168. [Online]. Available:
  \url{https://doi.org/10.1145/3173162.3173206}
\BIBentrySTDinterwordspacing

\bibitem{AdaptCap}
\BIBentryALTinterwordspacing
C.~Hankendi, A.~K. Coskun, and H.~Hoffmann, ``Adapt{\&}cap: Coordinating
  system- and application-level adaptation for power-constrained systems,''
  \emph{{IEEE} Des. Test}, vol.~33, no.~1, pp. 68--76, 2016. [Online].
  Available: \url{https://doi.org/10.1109/MDAT.2015.2463275}
\BIBentrySTDinterwordspacing

\bibitem{Givargis2001}
T.~Givargis, F.~Vahid, and J.~Henkel, ``System-level exploration for
  pareto-optimal configurations in parameterized systems-on-a-chip,'' in
  \emph{Computer Aided Design, 2001. ICCAD 2001. IEEE/ACM International
  Conference on}, Nov 2001, pp. 25--30.

\bibitem{Palermo2009}
G.~Palermo, C.~Silvano, and V.~Zaccaria, ``Respir: A response surface-based
  pareto iterative refinement for application-specific design space
  exploration,'' \emph{IEEE Transactions on Computer-Aided Design of Integrated
  Circuits and Systems}, vol.~28, no.~12, pp. 1816--1829, Dec 2009.

\bibitem{Zitzler2012}
E.~Zitzler, M.~Laumanns, and L.~Thiele, ``Spea2: Improving the strength pareto
  evolutionary algorithm,'' Tech. Rep., 2001.

\bibitem{Knowles1999}
J.~Knowles and D.~Corne, ``The pareto archived evolution strategy: a new
  baseline algorithm for pareto multiobjective optimisation,'' in
  \emph{Evolutionary Computation, 1999. CEC 99. Proceedings of the 1999
  Congress on}, vol.~1, 1999, p. 105 Vol. 1.

\bibitem{bienia11benchmarking2011}
C.~Bienia, ``Benchmarking modern multiprocessors,'' Ph.D. dissertation,
  Princeton University, January 2011.

\bibitem{Che2009}
S.~Che, M.~Boyer, J.~Meng, D.~Tarjan, J.~W. Sheaffer, S.~H. Lee, and
  K.~Skadron, ``Rodinia: A benchmark suite for heterogeneous computing,'' in
  \emph{Workload Characterization, 2009. IISWC 2009. IEEE International
  Symposium on}, Oct 2009, pp. 44--54.

\end{thebibliography}

\end{document}